\documentclass[lettersize,journal]{IEEEtran}
\usepackage{tikz,xcolor,hyperref}
\definecolor{lime}{HTML}{A6CE39}
\DeclareRobustCommand{\orcidicon}{%
    \begin{tikzpicture}
    \draw[lime, fill=lime] (0,0) 
    circle [radius=0.16] 
    node[white] {{\fontfamily{qag}\selectfont \tiny ID}};    \draw[white, fill=white] (-0.0625,0.095) 
    circle [radius=0.007];    \end{tikzpicture}
    \hspace{-2mm}}
\foreach \x in {A, ..., Z}{%
    \expandafter\xdef\csname orcid\x\endcsname{\noexpand\href{https://orcid.org/\csname orcidauthor\x\endcsname}{\noexpand\orcidicon}}
    }

\usepackage{amsmath,amsfonts}
\usepackage{algorithm}
\usepackage{algpseudocode}
\usepackage{array}

\usepackage{textcomp}
\usepackage{stfloats}
\usepackage{url}
\usepackage{verbatim}
\usepackage{graphicx}
\usepackage{cite}
\usepackage{amsmath}
\usepackage{amssymb}
\usepackage{mathtools}
\usepackage{amsthm}
\usepackage{hyperref}
\usepackage{url}
\usepackage[utf8]{inputenc} 
\usepackage[T1]{fontenc}    
\usepackage{hyperref}       
\usepackage{url}            
\usepackage{booktabs}       
\usepackage{amsfonts}       
\usepackage{nicefrac}       
\usepackage{microtype}      
\usepackage{xcolor}         
\usepackage{graphicx}
\usepackage{wrapfig}
\usepackage{multirow}
\usepackage{soul}
\usepackage{makecell}

\usepackage{subcaption}
\usepackage{comment}
\usepackage{threeparttable}
\usepackage{float}
\usepackage{bm}
\usepackage[english]{babel}
\def\BibTeX{{\rm B\kern-.05em{\sc i\kern-.025em b}\kern-.08em
    T\kern-.1667em\lower.7ex\hbox{E}\kern-.125emX}}
\usepackage{balance}
\begin{document}
\title{FAIM: Frequency-Aware Interactive Mamba for Time Series Classification}

\author{Da~Zhang\orcidA{},~\IEEEmembership{Student Member,~IEEE,} 
Bingyu~Li\orcidB{}, 
Zhiyuan~Zhao\orcidC{},
Yanhan Zhang,
Junyu~Gao\orcidC{},~\IEEEmembership{Member,~IEEE,} Feiping~Nie\orcidD{},~\IEEEmembership{Senior Member,~IEEE,}~and~Xuelong~Li\orcidE{},~\IEEEmembership{Fellow,~IEEE}

\thanks{Da Zhang, Junyu Gao, and Feiping Nie are with the School of Artificial Intelligence, OPtics and ElectroNics (iOPEN), Northwestern Polytechnical University, Xi'an 710072, China and also with the Institute of Artificial Intelligence (TeleAI), China Telecom, China. (E-mail: dazhang@mail.nwpu.edu.cn; gjy3035@gmail.com; feipingnie@gmail,com).}

\thanks{Bingyu Li, Zhiyuan Zhao, Yanhan Zhang, and Xuelong Li are with the Institute of Artificial Intelligence (TeleAI), China Telecom, China. (E-mail: libingyu0205@mail.ustc.edu.cn; tuzixini@gmail.com; zhangyh78@chinatelecom.cn;
xuelong\_li@ieee.org).}

\thanks{Junyu Gao and Xuelong Li are corresponding authors.}
}


\maketitle

\begin{abstract}
Time series classification (TSC) is crucial in numerous real-world applications, such as environmental monitoring, medical diagnosis, and posture recognition. TSC tasks require models to effectively capture discriminative information for accurate class identification. Although deep learning architectures excel at capturing temporal dependencies, they often suffer from high computational cost, sensitivity to noise perturbations, and susceptibility to overfitting on small-scale datasets. To address these challenges, we propose FAIM, a lightweight Frequency-Aware Interactive Mamba model. Specifically, we introduce an Adaptive Filtering Block (AFB) that leverages Fourier Transform to extract frequency-domain features from time series data. The AFB incorporates learnable adaptive thresholds to dynamically suppress noise and employs element-wise coupling of global and local semantic adaptive filtering, enabling in-depth modeling of the synergy among different frequency components. Furthermore, we design an Interactive Mamba Block (IMB) to facilitate efficient multi-granularity information interaction, balancing the extraction of fine-grained discriminative features and comprehensive global contextual information, thereby endowing FAIM with powerful and expressive representations for TSC tasks. Additionally, we incorporate a self-supervised pre-training mechanism to enhance FAIM's understanding of complex temporal patterns and improve its robustness across various domains and high-noise scenarios. Extensive experiments on multiple benchmarks demonstrate that FAIM consistently outperforms existing state-of-the-art (SOTA) methods, achieving a superior trade-off between accuracy and efficiency and exhibits outstanding performance. Our code is available at \href{https://github.com/zhangda1018/FAIM}{FAIM}.

\end{abstract}


\section{Introduction}
\label{sec:intro}

\IEEEPARstart{T}{ime} Series Classification (TSC) \cite{liu2025multiscale, wang2025sagog} plays a crucial role in a wide range of real-world scenarios, such as healthcare \cite{wang2024medformer}, human activity recognition \cite{li2023human}, and environmental monitoring \cite{russwurm2023end}. 
With the continuous deployment of automated devices, sequential observational data has grown dramatically both in scale and complexity, often exhibiting high dimensionality, multivariate structure, and strong noise interference \cite{eldele2024tslanet}. 
This places increasing demands on model discriminability and robust feature representation to adapt to diverse and complex real-world environments.

Traditional TSC methods, such as those based on symbolic representations \cite{li2022new} or subsequence shapelets \cite{le2024shapeformer}, often rely on cumbersome preprocessing or manual feature engineering. 
When facing multivariate or highly redundant data, these approaches can lead to bloated feature spaces and difficulties in flexible expansion. 
Recently, deep learning architectures such as RNNs, CNNs, MLPs, and Transformers have gradually become mainstream for TSC tasks \cite{mohammadi2024deep}. 
RNNs have the natural advantage of capturing temporal dependencies, but are limited by issues such as vanishing gradients, making it hard to model long-range dependencies. 
CNNs are effective in extracting local patterns but often ignore global semantic levels \cite{eldele2021time}. 
Transformers leverage self-attention mechanisms to balance local and global dependencies, significantly advancing TSC performance. 
However, their high computational cost and dependence on large-scale data present challenges such as heavy resource consumption, sensitivity to noise, and overfitting in small-sample scenarios \cite{Yuqietal-2023-PatchTST}. 
Some studies show that MLPs can surprisingly outperform Transformers on sequence tasks, but their simple linear structures make it difficult to handle complex and noisy data \cite{zeng2023transformers}.

\begin{figure}[t]
\centering
\includegraphics[width=0.98\columnwidth]{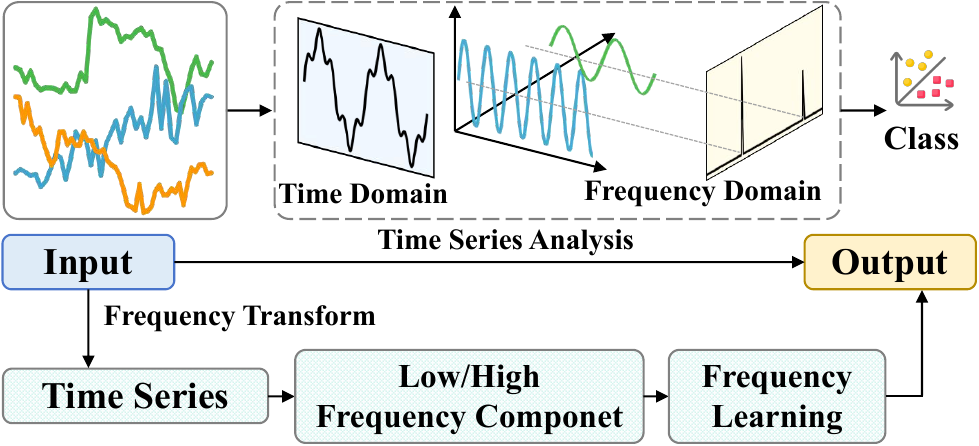} 
\caption{Pipeline of how frequency transform acts in TSC framework.}
\label{fig1}
\end{figure}

In recent years, State Space Models (SSMs) \cite{gu2022efficiently, gu2020hippo} have attracted increasing attention due to their linear complexity and ability to effectively model long-range dependencies, providing a new high-efficiency framework for temporal data representation. 
Exemplified by Mamba \cite{gu2023mamba}, SSM-based models utilize mechanisms such as selective scanning to enable more efficient serial and parallel processing of sequence information and enhanced modeling of long-range features \cite{tian2025before}.
However, these models still overlook feature extraction in the frequency domain (a pipeline through frequency transformation is provided in Figure \ref{fig1}) and the fusion of multi-granularity information, leaving ample room for improvement in generalization under noisy or limited-sample conditions \cite{ahamed2025tscmamba}.

To address these challenges, we propose FAIM, a lightweight Frequency-Aware Interactive Mamba model. 
The core of FAIM consists of an Adaptive Filtering Block (AFB), which decomposes the raw sequence data in the frequency domain via Fourier Transform and leverages learnable thresholds for dynamic noise suppression to highlight key signals. 
Inspired by the convolution theorem \cite{huang2023adaptive, han2025content}, AFB employs learnable global and local filters that automatically modulate various frequency components via element-wise multiplication, mimicking circular convolution to capture both short- and long-term interactions. 
The processed signals are then mapped back to the time domain through inverse Fourier Transform, yielding enhanced feature representations. 
Furthermore, FAIM introduces an Interactive Mamba Block (IMB), which promotes the interaction of multi-granularity discriminative information for temporal pattern extraction using a dual-causal convolution kernel structure, further improving the richness of learned representations. 
To strengthen robustness in complex scenarios, we incorporate a self-supervised learning mechanism, enabling the model to better exploit the intrinsic structure of sequential data. 
Extensive experiments on multiple benchmarks demonstrate that FAIM consistently outperforms existing state-of-the-art methods, achieving a superior balance between accuracy and efficiency. In summary, our main contributions are as follows:

\begin{itemize}
    \item We propose a novel lightweight frequency-aware interactive Mamba model for effective modeling of sequential data.
    \item Adaptive Filtering Block (AFB) that leverages learnable thresholds and a global–local coupling mechanism is designed to effectively suppress noise and model dynamic collaboration among frequency components.
    \item We introduce an Interactive Mamba Module (IMB) that employs dual-causal convolution structures to facilitate the interaction of multi-granularity temporal features. 
    \item Extensive experiments demonstrate that our method surpasses SOTA baselines, proving its effectiveness and superiority.
\end{itemize}

\section{Related Works}
\label{related_works}
\subsection{Time Series Frequency-Aware Analysis}
Frequency-aware analysis has become an important direction in time series research \cite{zhang2025beyond, murad2025wpmixer}. 
By leveraging frequency-domain information, it is possible to gain deeper insights into temporal data and enhance modeling capabilities \cite{liu2023temporal}. 
Early studies primarily relied on techniques such as FFT to transform time-domain signals into frequency components for extracting periodic patterns and suppressing noise \cite{wu2021autoformer}, thereby improving the robustness of downstream models. 
In recent years, the integration of frequency analysis and deep learning has achieved remarkable progress \cite{luo2025tfdnet}. 
For example, models such as FEDFormer \cite{zhou2022fedformer} and FiLM \cite{zhou2022film} utilize frequency information as additional features to enhance the capture of long-term periodic patterns and noise filtration; TimesNet \cite{wu2023timesnet} decomposes periodic features via FFT and further strengthens representation through high-frequency filtering, while FITS \cite{xu2024fits} directly trains sequences in the frequency domain. 
Hybrid architectures \cite{pang2024time, huang2025revisiting} that fuse time-domain and frequency-domain features have also been proposed, enabling the modeling of multi-scale dependencies and improved forecasting accuracy by combining information from both domains. 
However, existing methods typically rely on feature engineering for selecting periods and harmonics, which leads to issues such as overfitting and inefficiency. 
Inspired by frequency filters in computer vision \cite{chen2024frequency, yang2025ffr}, this work leverages learnable Fourier filters to promote mutual information learning, thereby enhancing semantic adaptability and reducing computational costs.

\subsection{Mamba in Time Series}
As an emerging sequential modeling architecture, Mamba \cite{gu2023mamba} is centered on a selective state space mechanism \cite{gu2022efficiently} that dynamically updates subsets of states at each time step, enabling it to effectively capture long-term dependencies. 
In contrast to traditional Transformers \cite{vaswani2017attention}, which face high computational complexity when dealing with long sequences, Mamba leverages structured state spaces to achieve linear time complexity and significantly improve processing efficiency \cite{wu2025affirm}.
In recent years, Mamba and its variants have been extensively explored for time series tasks. 
For example, MambaTS \cite{cai2024mambats} alleviates order sensitivity by scanning and rearranging historical information according to temporal variations. 
TimeMachine \cite{ahamed2024timemachine} introduces a multi-scale Mamba to address challenges in channel mixing and independence. 
DST-Mamba \cite{he2025decomposed} enables more comprehensive information interaction through a bidirectional block structure. 
Additionally, DPEM \cite{hou2025dpem} enhances the modeling of long-term dependencies by utilizing both forward and backward structures, while TSC-Mamba \cite{ahamed2025tscmamba} adapts to multi-timescale sequential data with multirate state spaces. 
In FAIM, we employ Mamba’s dual-causal convolutional structure to facilitate interactions among multi-granularity temporal features.

\begin{figure*}[t]
\centering
\includegraphics[width=1\linewidth]{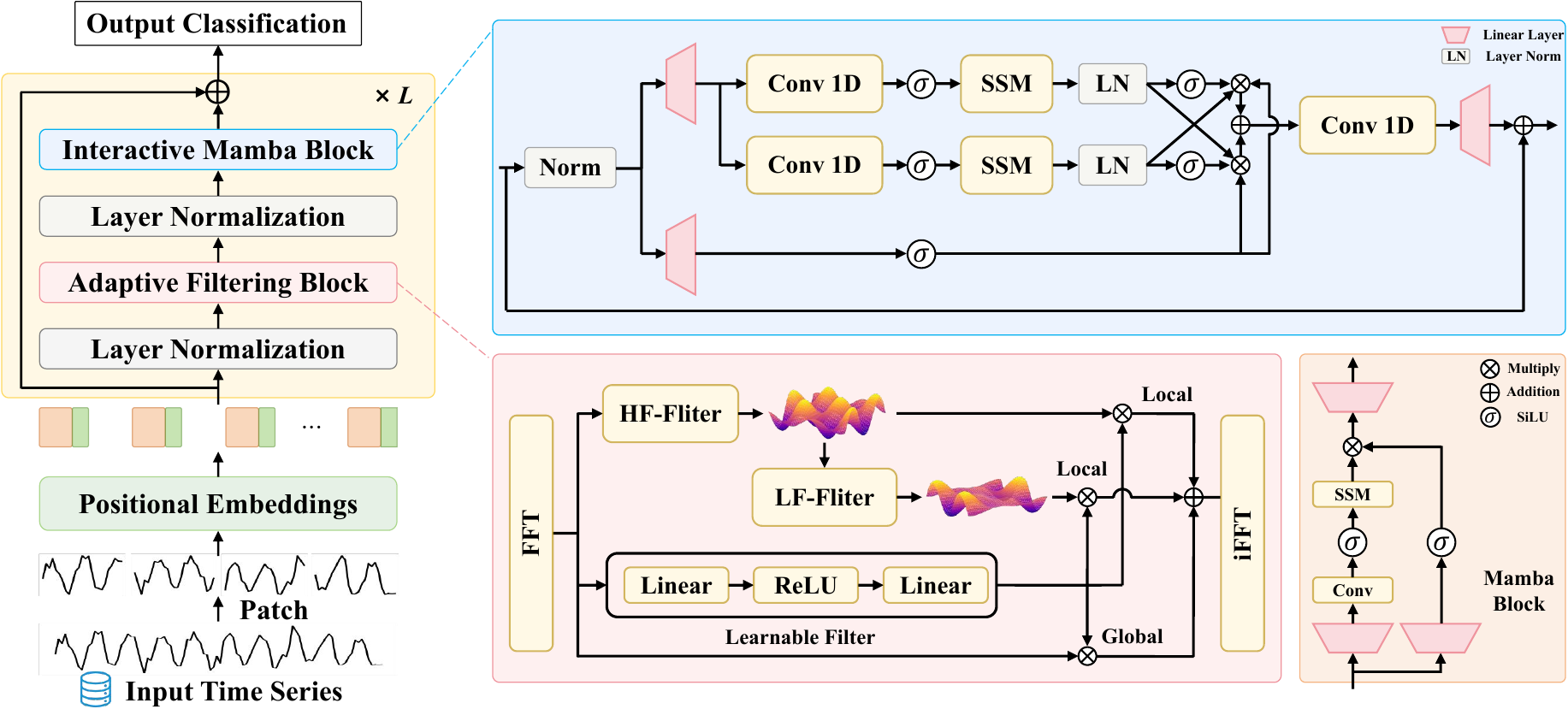} 
\caption{The structure of our proposed FAIM.
FAIM processes time series data with a hierarchical architecture consisting of multiple layers, each including an Adaptive Filtering Block (AFB), an Interactive Mamba Block (IMB), and Layer Normalization (LN). The model starts by dividing the input into patches and applying positional embeddings to retain temporal information. The AFB transforms the data to the frequency domain using a Fourier Transform, where high- and low-frequency noise is suppressed using adaptive and learnable filters. After iFFT reconstruction, the IMB further enhances the features via interactive convolution-activation, improving sensitivity to temporal dynamics. The final features are processed through the classification layer to obtain the final result.
}
\label{fig2}
\end{figure*}

\section{Proposed Methods}
\subsection{Preliminaries}

\subsubsection{Discrete Fourier Transform}
The Discrete Fourier Transform (DFT) is the foundation of our framework due to its dispersive properties are highly compatible with digital processing, and it can perform efficient numerical calculations with a complexity of $O(N \log N)$. 
The 1D DFT transforms an $N$-point complex sequence $x[n]$ (where $0 \leq n \leq N-1$) from the time domain to the frequency domain:
\begin{equation}
X[k] = \sum_{n=0}^{N-1} x[n] e^{-j(2\pi/N)kn} := \sum_{n=0}^{N-1} x[n] W_N^{kn},
\end{equation}
where $j$ is the imaginary unit and $W_N = e^{-j(2\pi/N)}$. The sequence $x[n]$ is represented in the frequency domain as $X[k]$ at frequency $\omega_k = 2\pi k / N$, which appears with a period of length $N$. Therefore, it is sufficient to consider the first $N$ points. Due to the bijective property of the DFT, the original sequence can be reconstructed via the inverse transform:
\begin{equation}
x[n] = \frac{1}{N} \sum_{k=0}^{N-1} X[k] e^{j(2\pi/N)kn}.
\end{equation}

\subsubsection{State Space Models}
SSMs project the continuous input signal $x(t)$ into the hidden state $h(t)$ and further map it to $y(t)$ to capture the temporal evolution of the signal states. This process is formulated as follows:
\begin{equation}
h'(t) = \mathbf{A}h(t) + \mathbf{B}x(t), y(t) = \mathbf{C}h(t),
\end{equation}
where $h'(t) = \frac{dh(t)}{dt}$, $\mathbf{A}$ denotes the state transition matrix, and $\mathbf{B}$ and $\mathbf{C}$ are projection matrices. Mamba belongs to discrete SSMs, and by synchronizing the time step $\Delta$ and applying the discretization rule, the continuous matrices $\mathbf{A}$ and $\mathbf{B}$ yield the discrete matrices $\overline{\mathbf{A}}$ and $\overline{\mathbf{B}}$:
\begin{equation}
\overline{\mathbf{A}} = \exp(\Delta\mathbf{A}), 
\overline{\mathbf{B}} = (\Delta\mathbf{A})^{-1} \left(\exp(\Delta\mathbf{A}) - \mathbf{I}\right) \cdot \Delta\mathbf{B},
\end{equation}
where $\mathbf{I}$ denotes identity matrix. Thus, SSMs can discretize continuous signals into sequences with time step size $\Delta$:
\begin{equation}
h_t = \overline{\mathbf{A}} h_{t-1} + \overline{\mathbf{B}} x_t, \quad y_t = \mathbf{C}h_t
\end{equation}
Finally, the model outputs the result via global convolution.

\subsection{Overview Framework}

An overview of FAIM is illustrated in Figure \ref{fig2}. 
The model input consists of a set of time series data (represented as a one-dimensional sequence in Figure \ref{fig2} for illustration purposes). 
To efficiently extract valuable features from various types of data, the model employs a scalable hierarchical architecture (with an adjustable number of layers $L$). Each layer contains an AFB, an IMB, and a Layer Normalization (LN) for stabilizing the training process. 
Specifically, the input first undergoes patching, followed by positional embedding to preserve temporal information in the sequence. 
Next, the AFB uses Fourier Transform to convert the time series data into the frequency domain, where adaptive thresholding attenuates high-frequency and low-frequency noise. Learnable High-Frequency (HF) and Low-Frequency (LF) filters are used to highlight relevant spectral features while reducing model size. 
Then, learnable filters (consisting of a linear layer, a ReLU function, and another linear layer) are applied to capture composite signals. 
After processing, the time-domain representation with reduced noise and enhanced features is reconstructed via iFFT. 
The IMB further refines these features through an interactive convolution-activation function, improving the model's ability to adapt to temporal dynamics in TSC. 
Together, these components effectively balance local and global temporal features, supporting robust time series analysis. The pseudocode for FAIM is provided in Algorithm~\ref{algorithm1}.

\begin{algorithm}[h]
\label{algorithm1}
\caption{FAIM: Frequency-Aware Interactive Mamba}
{Input:} Time series $\mathbf{X} \in \mathbb{R}^{N\times T}$ \\
{Output:} Classification result $\hat{\mathbf{y}}^{(c)}$
\begin{algorithmic}[1]
\State Divide $\mathbf{X}$ into patches $\{{B}_1, \dots, {B}_Z\}$
\State {// Embedding Layer}
\For{each patch ${B}_i \in \mathbb{R}^{N \times b}$}
    \State ${B}_i' \gets \mathrm{Conv1D}({B}_i)$ \Comment{Patch Embedding}
    \State ${B}_i^E \gets {B}_i' + {E}_i$ \Comment{Add Positional Encoding}
\EndFor
\For{$Layer = 1$ to $L$}
    \State {// Adaptive Filtering Block (AFB)}
    \State $\mathbf{F} = \mathcal{F}[B^E] \in \mathbb{C}^{N \times T'}$ \Comment{Fourier Transform}
    \State $\mathbf{F}_{Filter}^{{high}} = \mathbf{F} \odot (|\mathbf{M}| \leq \theta_{high})$
    \Comment{High-frequency Filtering}
    \State $\mathbf{F}_{Filter}^{{low}} = \mathbf{F} \odot (|\mathbf{M}| \geq \theta_{low})$ 
    \Comment{Low-frequency Filtering}
    \State $\mathbf{F}_G = \psi(\mathbf{F})_G \odot \mathbf{F}$ \Comment{Learnable Linear Filters}
    \State $  \mathbf{F}_L^{{low}} = \psi(\mathbf{F}^{{high}}_{{Filter}})_L \odot \mathbf{F}^{{low}}_{{Filter}}$ 
    \State $ \mathbf{F}_L^{{low}} = \psi(\mathbf{F}^{{low}}_{{Filter}})_L \odot \mathbf{F}^{{low}}_{{Filter}}$ 
    \State $\mathbf{F}_{integrated} = \mathbf{F}_G + \mathbf{F}_L^{{low}} + \mathbf{F}_L^{{low}}$ \Comment{Feature Integration}
    \State {// Interactive Mamba Block (IMB)}
    \State $\mathbf{H}_1 = \sigma(h_{t\_1}) \odot h_{t\_2} \odot (Linear(\mathbf{F}_T))$
    \State $\mathbf{H}_2 = \sigma(h_{t\_2}) \odot h_{t\_1} \odot (Linear(\mathbf{F}_T))$
    \State $\mathbf{H}_{IMB} = Conv\_3 (\mathbf{H}_1 + \mathbf{H}_2)$ \Comment{Layer output}
\EndFor
\State {// Classification decoding}
\State $\hat{\mathbf{y}}^{(c)} \gets \mathrm{ClsHead}(\mathbf{H}_{IMB})$
\State \Return $\hat{\mathbf{y}}^{(c)}$
\end{algorithmic}
\label{algorithm1}
\end{algorithm}

\subsection{Embedding Layer}

Given an input dataset $\mathcal{D}$, where each sample $\mathbf{X} \in \mathbb{R}^{N \times T}$ has $N$ dimensions and a sequence length $T$. First, the sample $\mathbf{X}$ is divided into a set of $Z$ patches $\{B_1, B_2, \ldots, B_Z\}$ and the dimension of each patch is determined by a predefined patch size $b$, so each patch $B_i \in \mathbb{R}^{N \times b}$. Then, each patch is mapped to a new dimension $b'$ through a one-dimensional convolutional layer, resulting in $B_i \stackrel{Conv1D}{\longrightarrow}  B_i' \in \mathbb{R}^{N \times b'}$.
Next, position embeddings $E_i$ are added to preserve the temporal order disrupted during segmentation. The final position embedding obtained is
$ B^E_i = B_i' + E_i$. Thus, we obtain $B^E = \{B^E_1, B^E_2, \ldots, B^E_Z\}$, and these embedding vectors are learnable.

\subsection{Adaptive Filtering Block}

For time series from different domains, information of various types is often contained in different frequency components. 
Conventional denoising methods typically rely heavily on low-frequency components. 
This leads to two main problems: 
(1) useful high-frequency details may be filtered out indiscriminately; and (2) some low-frequency noise might also be retained in the output, resulting in reduced signal-to-noise ratio and increased error rate \cite{wu2025affirm}. 
To address this, we propose a dynamic filtering module based on global circular convolution, designed to adapt to temporal features in various types of signals.

\subsubsection{Fast Fourier Transform}

Given $B^E$, the FFT is applied to obtain the frequency domain representation $\mathbf{F}$:
\begin{equation}
    \mathbf{F} = \mathcal{F}[B^E] \in \mathbb{C}^{N \times T'},
\end{equation}
where $\mathcal{F}[\cdot]$ denotes FFT operation, and $T'$ is the length of the sequence in the frequency domain, which generally differs from $T$ and depends on the FFT implementation and dataset characteristics. Each variable in time series is transformed independently, yielding a comprehensive frequency-domain representation of all variables' original time series.

\subsubsection{High-frequency Filtering}

High-frequency components typically reflect apparent noise, abrupt changes, or rapid oscillations. Therefore, removing high-frequency noise helps the model to learn underlying trends and periodic patterns. By setting a trainable threshold $\theta_{high}$, the network adaptively selects and retains only the frequencies lower than this threshold while suppressing most high-frequency noise, thus preserving important low-frequency trends. Specifically,
\begin{equation}
\mathbf{F}_{Filter}^{{high}} = \mathbf{F} \odot (|\mathbf{M}| \leq \theta_{high}),
\end{equation}
where $\odot$ denotes the Hadamard product and $|\mathbf{M}| \leq \theta_{high}$ is a binary mask, retaining the components where frequency is lower than $\theta_{high}$, and masking out the rest.

\subsubsection{Low-frequency Filtering}

Low-frequency components often contain the underlying trends and periodicities of the time series data. However, due to systematic errors in data collection or processing, they may also contain residual artifacts. Therefore, after processing high-frequency noise, we introduce a low-frequency denoising branch. By setting a learnable parameter $\theta_{low}$, low-frequency components below this threshold are filtered out, enhancing the model's ability to handle non-smooth data:
\begin{equation}
\mathbf{F}_{Filter}^{{low}} = \mathbf{F} \odot (|\mathbf{M}| \geq \theta_{low}),
\end{equation}
where $|\mathbf{M}| \geq \theta_{low}$ indicates the mask that retains components with frequencies higher than $\theta_{low}$.

\subsubsection{Learnable Linear Filters}

We further introduce structured learnable linear filters to perform linear transformations on features at different frequency bands, allowing for a more detailed and expressive spectral representation. This process consists of three parts: global filtering of the original frequency $\mathbf{F}$ via $\psi(\mathbf{F})_G$, as well as local filtering for high and low-frequency components $\psi(\mathbf{F}^{{high}}_{{Filter}})_L$ and $\psi(\mathbf{F}^{{low}}_{{Filter}})_L$:
\begin{align}
    \mathbf{F}_G &= \psi(\mathbf{F})_G \odot \mathbf{F}, \\
    \mathbf{F}_L^{{low}} &= \psi(\mathbf{F}^{{high}}_{{Filter}})_L \odot \mathbf{F}^{{low}}_{{Filter}}, \\
    \mathbf{F}_L^{{low}} &= \psi(\mathbf{F}^{{low}}_{{Filter}})_L \odot \mathbf{F}^{{low}}_{{Filter}}.
\end{align}
As shown in Figure \ref{fig2}, $\psi(\cdot)$ consists of a linear layer, a ReLU activation, and another linear layer, enabling the network to achieve light structure. The elementwise operation in this step is equivalent to circular convolution (see proof 1 in the appendix), endowing the sequence with a larger receptive field and enabling better capture of periodic patterns in time series data. Next, we combine the output features of all filters to obtain a complete set of spectral features:
\begin{equation}
    \mathbf{F}_{integrated} = \mathbf{F}_G + \mathbf{F}_L^{{low}} + \mathbf{F}_L^{{low}},
\end{equation}

\subsubsection{Inverse Fourier Transform}

To convert the integrated frequency-domain features back to the time domain, we apply iFFT. The obtained time-domain signal $\mathbf{F}_T$ is:
\begin{equation}
    \mathbf{F}_T = \mathcal{F}^{-1}[\mathbf{F}_{integrated}] \in \mathbb{C}^{N \times T'},
\end{equation}
ensuring that the enhanced features are aligned with the original temporal structure of the input sequence.

\subsection{Interactive Mamba Block}

Building on the original Mamba architecture, we propose the IMB with causal convolution, equipped with convolution kernels of different sizes to capture local features and long-range dependencies within a broader range. 
Specifically, the first convolution uses a smaller $2 \times 2$ kernel to capture fine-grained local patterns, whereas the second utilizes a larger $4 \times 4$ kernel to identify broader dependencies. 
The unified formulation is expressed as follows:
\begin{equation}
    h_{t\_i} = LN \left( SSM \left( \sigma \left( Conv\_i (Linear(\mathbf{F}_T)) \right) \right) \right),
\end{equation}
where $Conv\_i$ ($i=1,2$) denotes one-dimensional convolution, $\sigma$ is the SiLU activation function, and LN denotes layer normalization. 
IMB allows the output of each layer to modulate the feature extraction of the other, and uses elementwise product to realize cross-interaction between features extracted at different scales, which facilitates better modeling of complex relationships:
\begin{align}
    \mathbf{H}_1 &= \sigma(h_{t\_1}) \odot h_{t\_2} \odot (Linear(\mathbf{F}_T)), \\
    \mathbf{H}_2 &= \sigma(h_{t\_2}) \odot h_{t\_1} \odot (Linear(\mathbf{F}_T)),
\end{align}
Then, $\mathbf{H}_1$ and $\mathbf{H}_2$ are concatenated and passed through a final linear convolutional layer $Conv\_3$ to obtain the enhanced features for classification:
\begin{equation}
    \mathbf{H}_{IMB} = Conv\_3 (\mathbf{H}_1 + \mathbf{H}_2).
\end{equation}

\subsection{Self-supervised Pre-training}

Inspired by PatchTST \cite{Yuqietal-2023-PatchTST}, we introduce a self-supervised pre-training scheme. 
This approach uses a masked autoencoder paradigm to process sequential time series signals, enabling the model to learn high-level representation.
Specifically, the entire process is divided into pre-training and formal training phases. 
The pseudocode is provided in Algorithm~\ref{algorithm2}.

\subsubsection{Pre-training Stage}
During the pre-training stage, an adaptive threshold is used to select masks for the input sequence, and FAIM accurately reconstructs these masked segments. 
This method forces the model to learn useful feature representations during the reconstruction task, inferring the interdependencies within the data. 
We use the Mean Squared Error (MSE) loss function \( \mathcal{L}_{MSE} \) to optimize the model's ability, helping the model understand the intrinsic structure of the data.
\begin{equation}
\mathcal{L}_{MSE} = \frac{1}{\sum_{i=1}^{T} \lambda_i} \sum_{i=1}^{T} \lambda_i \cdot (x_i - \hat{x}_i)^2,
\end{equation}
\noindent where \( T \) is the length of the sequence, \( x_i \) is the true value of the \( i \)-th time point, \( \hat{x}_i \) is the predicted value of the \( i \)-th time point, and \( \lambda_i \) is the mask value, determining whether to calculate the error for that time point.

\subsubsection{Formal Training Stage}
After pre-training, we use the optimal weights to initialize the model and conduct supervised training for TSC. We use the label-smooth cross-entropy loss function \( \mathcal{L}_{SCE} \) for optimization. 
By introducing label smoothing, we fine-tune the complete data, which helps improve the model's robustness and enhance its performance.
\begin{equation}
\mathcal{L}_{SCE} = -\sum_{i=1}^{k} y_i^{\text{smooth}} \cdot \log(\hat{y}_i),
\end{equation}
\begin{equation}
y_i^{\text{smooth}} = (1 - \epsilon) \cdot y_i + \frac{\epsilon}{k},
\end{equation}
where \( y_i^{\text{smooth}} \) is the true class label of the one-hot encoded label after label smoothing, \( \hat{y}_i \) is the predicted probability of each class, \( k \) is the number of classes, and label smoothing adjusts the true label's smooth parameter \( \epsilon \) (usually a small value) to prevent the model from overconfidence and enhance its generalization ability.

\begin{algorithm}[t]
\caption{FAIM Training Strategy}
{Input:} Training samples $\mathbf{X}$, label $\mathbf{y}$ \\
{Output:} Trained model parameters $\Theta$
\begin{algorithmic}[1]
\State {Pre-training:}
\While{not converged}
    \State Randomly mask a portion of input $\mathbf{X}$
    \State Obtain reconstructed output $\hat{\mathbf{X}} = \text{FAIM}(\mathbf{X}_{\text{masked}})$
    \State Compute loss $\mathcal{L}_{\mathrm{MSE}} = \text{MSE}(\hat{\mathbf{X}}, \mathbf{X}_{\text{true}})$
    \State Update parameters $\Theta$ using $\mathcal{L}_{\mathrm{MSE}}$
\EndWhile
\State {Fine-tuning:}
\While{not converged}
    \State Get output $(\hat{\mathbf{y}}^{(c)}, \hat{\mathbf{X}}^{(f)}, \hat{\mathbf{a}}^{(a)}) = \text{FAIM}(\mathbf{X})$
        \State Compute $\mathcal{L}_{\mathrm{SCE}} = \text{LabelSmoothedCE}(\hat{\mathbf{y}}^{(c)}, \mathbf{y})$
    \State Update parameters $\Theta$ using corresponding loss
\EndWhile
\State \Return $\Theta$
\end{algorithmic}
\label{algorithm2}
\end{algorithm}

\section{Experiments}
\subsection{Experimental Setup}
\subsubsection{Datasets and Evaluation Metrics}
Following \cite{eldele2024tslanet}, we conduct experiments on 85 univariate datasets from the UCR archive \cite{dau2019ucr} and 26 multivariate datasets from the UEA archive \cite{bagnall2018uea}. 
For selected UCR datasets which encompasses \textit{Image, Device, Motion, Sensor, Simulated, Spectro, and ECG} data types, the length range from 24 to 2709, the training sizes from 16 to 8926, and the number of classes varies from 2 to 60.
For selected UEA datasets, the length range from 15 to 20000, the training sizes from 12 to 30000, the number of classes varies from 2 to 39, and the dimension are from 2 to 963.
Detailed information is provided in the appendix. Following \cite{wang2025sagog}, we adopt widely used evaluation metrics for time series classification models, including the number of datasets achieving the highest accuracy (Num. Top-1), average accuracy (Avg. ACC), F1-Score, average rank (Avg. Rank), and critical difference (CD) for statistical significance testing.

\subsubsection{Baselines and Implementation Details}
We compare FAIM against ten baseline methods, namely MPTSTNet \cite{mu2025mptsnet}, FreRA \cite{tian2025frera}, TSLANet \cite{eldele2024tslanet}, GPT4TS \cite{zhou2023one}, TimesNet \cite{wu2023timesnet}, PatchTST \cite{Yuqietal-2023-PatchTST}, Rocket \cite{dempster2020rocket}, MLP \cite{zeng2023transformers}, TS2Vec \cite{yue2022ts2vec}, and TS-TCC \cite{eldele2021time}. 
All baseline models are trained under the same experimental settings as our method for a fair comparison.  
The experiments are conducted on a single NVIDIA A800 80GB GPU using PyTorch. The AdamW optimizer is adopted, with an initial learning rate of 1e-3 and a weight decay of 1e-4. The maximum batch size is set to 256 for both pre-training and fine-tuning stages. The pre-training stage runs for 100 epochs, while fine-tuning is conducted for 300 epochs.

\subsection{Classification Results}

\begin{table*}
\centering
\resizebox{1.0\linewidth}{!}{
\begin{tabular}{ll|ccccccccccc}
\toprule
\multicolumn{2}{c}{\textbf{Method}} & \textbf{FAIM} & \textbf{MPTSNet} & \textbf{FreRA} & \textbf{TSLANet} & \textbf{GPT4TS} & \textbf{TimesNet} & \textbf{PatchTST} & \textbf{Rocket} & \textbf{MLP} & \textbf{TS2Vec} & \textbf{TS-TCC} \\
\midrule
\multirow{3}{*}{\shortstack[l]{85 UCR }} 
& Avg. Acc      &  \textbf{0.849} & 0.745 & 0.830 & 0.831 & 0.612 & 0.652 & 0.718 & 0.814 & 0.696 & 0.814 &0.750      \\
& Avg. Rank     & \textbf{3.059} & 6.541 & 3.524 & 3.324 & 9.024 & 8.600 & 7.688 & 4.376 & 8.106 & 5.076 &6.682\\
& Num. Top-1    & \textbf{35}    & 3     & 25     & 16    & 2     & 0     & 3     & 20     & 2   & 10 & 4\\
\midrule
\multirow{3}{*}{\shortstack[l]{26 UEA }} 
& Avg. Acc      & \textbf{0.767} & 0.730 & 0.735 & 0.727 & 0.585 & 0.680 & 0.703 & 0.688 & 0.658 & 0.671 &0.653   \\
& Avg. Rank     & \textbf{2.885} & 5.000 & 4.712 & 3.769 & 8.673 & 7.154 & 5.308 & 6.231 & 7.481 & 6.808  & 7.981 \\
& Num. Top-1   & \textbf{9}     & 6 &  6     & 4     & 0     & 1     & 2     & 4     & 1     & 0  & 1 \\
\bottomrule
\end{tabular}
}
\caption{Classification comparison with SOTA methods.}
\label{table1}
\end{table*}

\begin{table*}[hbpt]
\centering
\resizebox{1.0\linewidth}{!}{
\begin{tabular}{c|c|c|c|c|c|c|c|c|c|c|c}
\toprule
Type & {FAIM (Ours)} & MPTSNet & FreRA & TSLANet & GPT4TS & TimesNet & PatchTST & Rocket & MLP & TS2Vec & TS-TCC \\
\midrule
Image  &0.839 	&0.748 	&0.832 	&\textbf{0.842} 	&0.602 	&0.658 	&0.717 	&0.821 	&0.734 	&0.815 	&0.746  \\
Motion &\textbf{0.798} 	&0.647 	&0.779 	&0.759 	&0.480 	&0.611 	&0.573 	&0.773 	&0.532 	&0.730 	&0.694  \\
ECG  &0.950 	&0.888 	&0.949 	&0.938 	&0.851 	&0.836  &0.863 	&\textbf{0.951}	&0.921 	&0.940 	&0.868 \\
Device & \textbf{0.706} 	&0.611 	&0.690 	&0.658 	&0.409 	&0.502 	&0.599 	&0.614 	&0.447 	&0.629 	&0.619 \\
Spectro &\textbf{0.898} 	&0.752 	&0.831 &0.812 	&0.652 	&0.565 	&0.850 	&0.828 	&0.812 	&0.875 	&0.634 \\
Simulated  &\textbf{0.979} 	&0.875 	&0.991 	&0.964 	&0.861 	&0.788 	&0.837 &0.898 	&0.812 	&0.928 	&0.887 \\
Sensor   &\textbf{ 0.865} 	&0.775 &	0.829 	&0.863 	&0.634 &	0.661 &	0.740 &	0.826 	&0.695 &	0.842 	&0.812        \\
\midrule
Average  &\textbf{0.849} 	&0.745 	&0.830 	&0.831 	&0.612 	&0.652 	&0.718 	&0.814 	&0.696 	&0.814 	&0.750 
\\
\bottomrule
\end{tabular}
}
\caption{The average accuracy of each category on 85 UCR datasets.}
\label{tab_2}
\end{table*}

\begin{table*}[hbpt]
\centering
\resizebox{1.0\linewidth}{!}{
\begin{tabular}{c|c|c|c|c|c|c|c|c|c|c|c}
\toprule
Dataset & {FAIM (Ours)} & MPTSNet & FreRA & TSLANet & GPT4TS & TimesNet & PatchTST & Rocket & MLP & TS2Vec & TS-TCC \\
\midrule
ArticularyWordRecognition & \textbf{0.993} & 0.973 & 0.990 & 0.990 & 0.933 & 0.962 & 0.977 & \textbf{0.993} & 0.973 & 0.987 & 0.953 \\
AtrialFibrillation & 0.533 & \textbf{0.600} & 0.467 & 0.400 & 0.333 & 0.333 & 0.533 & 0.200 & 0.466 & 0.200 & 0.267 \\
BasicMotions & {1.000} & 0.975 & {1.000} & {1.000} & 0.950 & 0.850 & 0.925 & {1.000} & 0.850 & 0.975 & {1.000} \\
Cricket & \textbf{1.000} & 0.888 & \textbf{1.000} & 0.986 & 0.083 & 0.875 & 0.847 & 0.986 & 0.916 & 0.972 & 0.917 \\
Epilepsy & 0.986 & 0.905 & \textbf{0.993} & 0.986 & 0.855 & 0.781 & 0.659 & 0.985 & 0.601 & 0.964 & 0.957 \\
EthanolConcentration & 0.395 & \textbf{0.425} & 0.323 & 0.304 & 0.254 & 0.277 & 0.289 & \textbf{0.425} & 0.334 & 0.308 & 0.285 \\
FaceDetection & 0.648 & \textbf{0.691} & 0.581 & 0.668 & 0.655 & 0.675 & 0.690 & 0.647 & 0.674 & 0.501 & 0.544 \\
FingerMovements & 0.630 & 0.620 & 0.610 & 0.610 & 0.570 & 0.594 & 0.620 & 0.610 & \textbf{0.640} & 0.480 & 0.460 \\
HandMovementDirection & 0.567 & 0.540 & 0.514 & 0.527 & 0.189 & 0.500 & \textbf{0.581} & 0.500 & 0.581 & 0.338 & 0.243 \\
Handwriting & 0.528 & 0.307 & \textbf{0.593} & 0.579 & 0.037 & 0.262 & 0.260 & 0.484 & 0.224 & 0.515 & 0.498 \\
Heartbeat & \textbf{0.819} & 0.746 & 0.785 & 0.776 & 0.365 & 0.745 & 0.766 & 0.697 & 0.731 & 0.480 & 0.764 \\
InsectWingbeat & 0.653 & \textbf{0.662} & 0.462 & 0.100 & 0.100 & 0.469 & 0.466 & 0.100 & 0.466 & 0.264 & 0.161 \\
JapaneseVowels & 0.989 & 0.964 & 0.965 & \textbf{0.992} & 0.981 & 0.978 & 0.987 & 0.956 & 0.978 & 0.984 & 0.980 \\
Libras & \textbf{0.933} & 0.838 & 0.911 & 0.928 & 0.794 & 0.778 & 0.811 & 0.838 & 0.733 & 0.867 & 0.772 \\
LSST & 0.615 & 0.599 & 0.494 & 0.663 & 0.463 & 0.592 & \textbf{0.678} & 0.541 & 0.357 & 0.537 & 0.447 \\
MotorImagery & \textbf{0.630} & 0.610 & 0.550 & 0.620 & 0.500 & 0.510 & 0.610 & 0.530 & 0.610 & 0.510 & 0.610 \\
NATOPS & \textbf{0.994} & 0.933 & 0.900 & 0.956 & 0.916 & 0.818 & 0.969 & 0.833 & 0.938 & 0.928 & 0.822 \\
PEMS-SF & 0.861 & \textbf{0.924} & 0.746 & 0.836 & 0.872 & 0.821 & 0.751 & 0.732 & 0.888 & 0.841 & 0.909 \\
PenDigits & \textbf{0.990} & 0.979 & 0.973 & 0.989 & 0.977 & 0.982 & 0.992 & 0.973 & 0.929 & 0.989 & 0.974 \\
PhonemeSpectra & 0.240 & 0.123 & \textbf{0.274} & 0.178 & 0.030 & 0.182 & 0.117 & 0.176 & 0.070 & 0.233 & 0.252 \\
RacketSports & 0.848 & 0.855 & 0.888 & \textbf{0.908} & 0.769 & 0.826 & 0.790 & 0.861 & 0.789 & 0.855 & 0.816 \\
SelfRegulationSCP1 & 0.887 & \textbf{0.928} & 0.908 & 0.918 & 0.914 & 0.774 & 0.898 & 0.884 & 0.844 & 0.874 & 0.886 \\
SelfRegulationSCP2 & 0.605 & \textbf{0.572} & \textbf{0.622} & 0.617 & 0.516 & 0.528 & 0.544 & 0.544 & 0.516 & 0.578 & 0.598 \\
SpokenArabicDigits & 0.998 & 0.990 & 0.984 & \textbf{0.999} & 0.993 & 0.990 & 0.997 & 0.992 & 0.968 & 0.932 & 0.973 \\
StandWalkJump & \textbf{0.667} & 0.466 & 0.667 & 0.467 & 0.333 & 0.533 & 0.600 & 0.467 & 0.600 & 0.467 & 0.533 \\
UWaveGestureLibrary & 0.931 & 0.856 & 0.900 & 0.913 & 0.843 & 0.833 & 0.800 & \textbf{0.944} & 0.818 & 0.884 & 0.753 \\
\midrule
Average & \textbf{0.767} & 0.730 & 0.735 & 0.727 & 0.585 & 0.680 & 0.703 & 0.688 & 0.658 & 0.671 & 0.653 \\
\midrule
1st count & \textbf{9} & 6 & 6 & 4 & 0 & 1 & 2 & 4 & 1 & 0 & 1 \\
\bottomrule
\end{tabular}
}
\caption{Classification accuracy on 26 UEA datasets.}
\label{tab_3}
\end{table*}

\begin{figure*}[ht]
    \begin{subfigure}[b]{0.48\linewidth}
        \includegraphics[width=\linewidth]{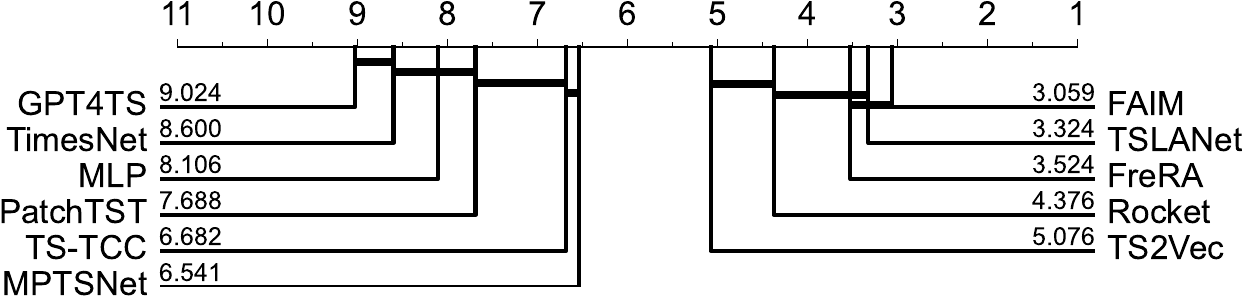}
        \label{cd_fig:sub1}
    \end{subfigure}
    \hfill
    \begin{subfigure}[b]{0.48\linewidth}
        \includegraphics[width=\linewidth]{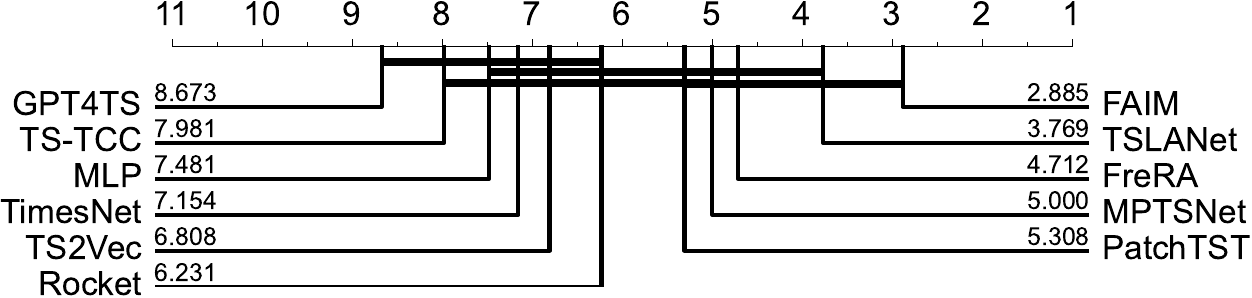}
        \label{cd_fig:sub2}
    \end{subfigure}
    \caption{CD diagram of SOTA methods on UCR (left) and UEA (right) datasets with a confidence level of 95\%.}
    \label{fig3}
\end{figure*}

Table~\ref{table1} provides a detailed comparison of the performance of FAIM and several mainstream time series classification methods. The results show that FAIM significantly outperforms the baselines on all evaluation metrics. Specifically, on the UCR dataset, FAIM achieved an average accuracy of 0.849, surpassing the second-best models TSLANet (0.831) and FreRA (0.830), which highlights its superior performance on a wide range of univariate time series datasets. On the UCR and UEA datasets, FAIM achieved the highest average accuracy of 0.849 and 0.765, respectively, and the lowest average ranks of 3.059 and 2.885, demonstrating its strong generalization ability and robustness on diverse time series data. FAIM also leads by a large margin in the number of datasets where it achieves the best accuracy (35 on UCR and 9 on UEA), which further confirms its superiority across different domains and under noisy conditions. This dominance in the number of datasets won is particularly noteworthy, suggesting that FAIM's frequency-aware and interactive Mamba components are highly effective at capturing the unique characteristics of various data distributions. Figure~\ref{fig3} shows the CD diagram for all datasets. The CD diagram clearly indicates that FAIM is statistically superior to most evaluated baselines, with a clear distinction in average rank from other models. For instance, the significant performance gap between FAIM and methods like GPT4TS and TimesNet underscores the limitations of purely attention-based or decomposition models in handling time series classification challenges, such as noise and computational cost, which FAIM’s design effectively mitigates.

To further investigate the model's performance on different data modalities, we present the average accuracy of each method on the UCR datasets, categorized by type, in Table \ref{tab_2}. The results show that FAIM achieves the highest accuracy in six out of seven categories, including complex data types like Image, Sensor, and ECG. This consistent advantage across multiple domains highlights FAIM's strong generalization ability and the robustness of its frequency-aware architecture in handling diverse time-series characteristics. The detailed results of all methods on each dataset are provided in the appendix.
Furthermore, to validate FAIM's effectiveness on multivariate time series, we conduct experiments on 26 UEA datasets, with detailed results shown in Table \ref{tab_3}. On these more complex multivariate datasets, FAIM again demonstrates its superior capability, achieving the highest average accuracy of 0.767 and securing the top rank on 9 datasets, more than any other competing method. This strong performance further confirms the model's robustness and its ability to effectively capture dependencies across multiple channels.

\subsection{Model Analysis}
\subsubsection{Various Variants of FAIM}

Table~\ref{table2} presents the results of ablation studies, evaluating the impact of removing key components from FAIM on task performance. 
Removing the AFB (i.e., w/o AFB) and IMB (i.e., w/o IMB) significantly degrades performance, with the absence of AFB causing a more pronounced decline. 
Specifically, for the NATOPS dataset, removing IMB leads to a decrease of 1.7\% in Acc and 1.8\% in F1. 
The removal of AFB results in a larger drop, by 4.5\% and 4.7\%, indicating that AFB plays a crucial role in feature extraction and denoising. 
Furthermore, we analyze the contributions of different submodules within AFB, namely high-frequency filtering (HF) and low-frequency filtering (LF), to noise suppression. 
Removing either filter (especially HF) has a significant negative effect on performance, suggesting that high-frequency noise has a greater impact on the results. 
When only global filtering is used (i.e., w/o HF+LF), the performance drops directly, confirming the critical role of local filters in noise management. 
In addition, the importance of pre-training is verified, as the lack of pre-training slightly reduces the model’s performance.

\begin{table}[h]
    \centering
    \resizebox{1.0\linewidth}{!}{
    \begin{tabular}{l|cc|cc|cc}
        \toprule
        \textbf{Variant} 
        & \multicolumn{2}{c|}{\textbf{Ham (UCR)}} 
        & \multicolumn{2}{c|}{\textbf{Heartbeat (UEA)}} 
        & \multicolumn{2}{c}{\textbf{NATOPS (UEA)}} \\
        & Acc & F1-Score & Acc & F1-Score & Acc & F1-Score  \\
        \midrule
w/o AFB      & 0.657 & 0.430 & 0.663 & 0.465 & 0.949 & 0.946\\
w/o HF       & 0.704 & 0.460 & 0.717 & 0.570 & 0.961 & 0.957\\
w/o LF       & 0.714 & 0.461 & 0.765 & 0.585 & 0.966 & 0.962\\
w/o HF+LF    & 0.666 & 0.441 & 0.751 & 0.583 & 0.955 & 0.951\\
w/o IMB      & 0.742 & 0.470 & 0.770 & 0.612 & 0.977 & 0.975\\
w/o Pretrain & 0.819 & 0.515 & 0.795 & 0.618 & 0.988 & 0.988\\
\midrule
FAIM         & 0.828 & 0.521 & 0.819 & 0.629 & 0.994 & 0.993 \\
        \bottomrule
    \end{tabular}
    }
    \caption{Ablation study of each component in FAIM.}
    
    \label{table2}
\end{table}

\subsubsection{Efficiency of Adaptive Filtering}

\begin{figure}[t]
    \includegraphics[width=0.468\textwidth]{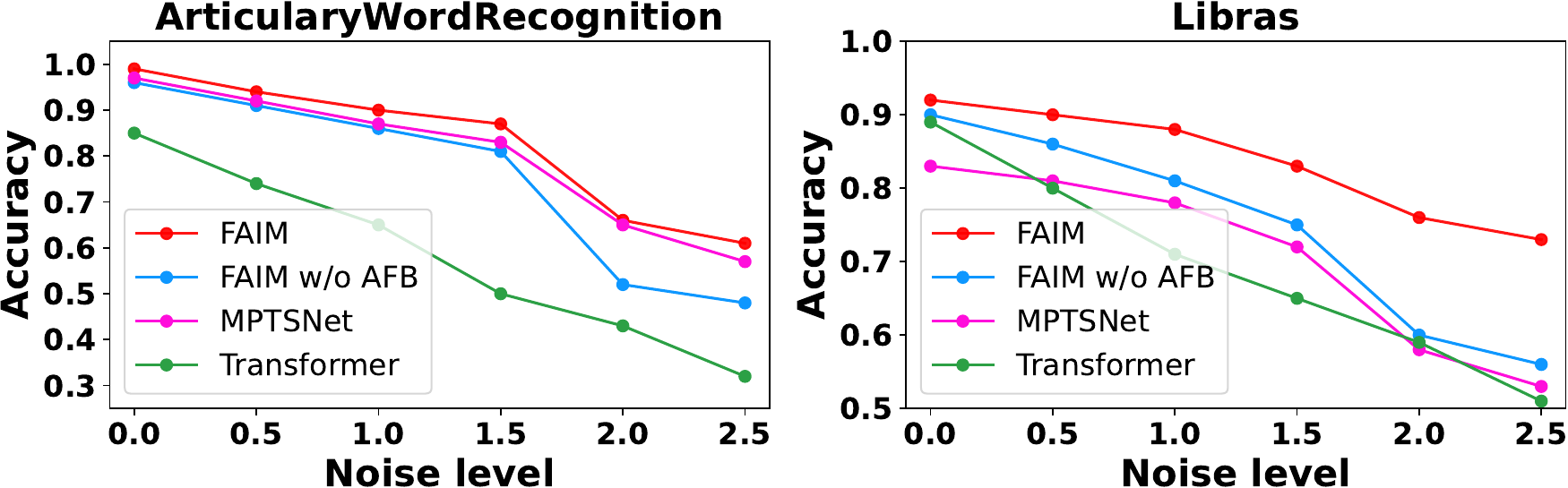}
    \caption{
    Robustness to noise levels on two UEA datasets: ArticularyWordRecognition and Libras.
    }
    \label{fig_noise}
\end{figure}

We conducted an in-depth investigation into the effectiveness of adaptive filters in mitigating noise and enhancing model robustness. 
Specifically, Figure~\ref{fig_noise} illustrates the performance of FAIM, compared with MPTSNet and Transformer models, with and without adaptive filters, after adding different levels of Gaussian noise to the time series.
As the noise level increases, the performance of MPTSNet and Transformer models drops rapidly. 
In contrast, FAIM maintains relatively stable performance, and its variants equipped with adaptive filters demonstrate the best noise resistance. 
Notably, when the noise level is high, the accuracy of the Transformer model drops sharply, while the decrease for FAIM with adaptive filters is much smaller.

\subsubsection{Scaling Efficiency}

\begin{figure}[ht]
    \begin{subfigure}[b]{0.49\linewidth}
        \includegraphics[width=\linewidth]{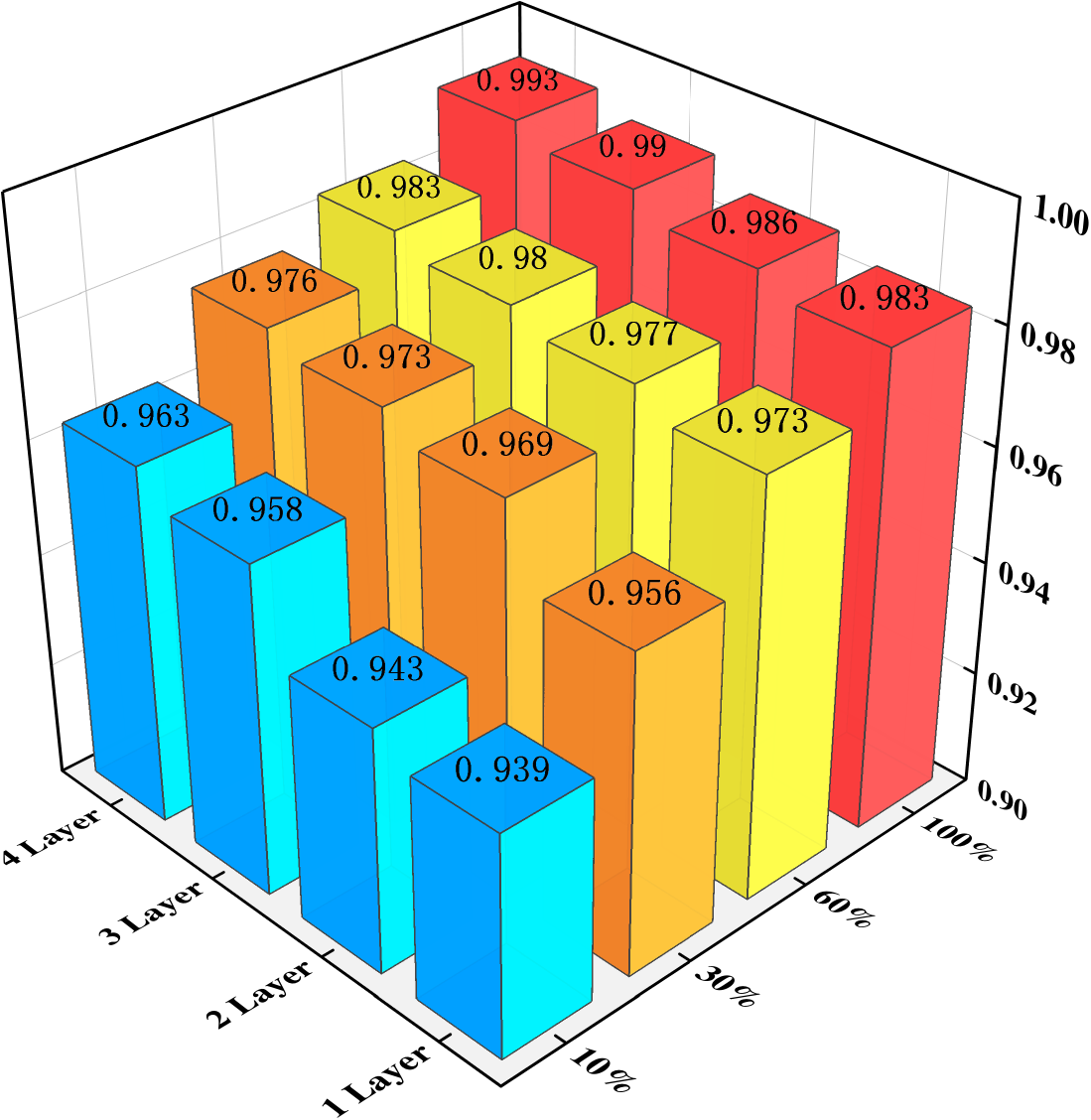}
        \caption{ArticularyWordRecognition}

    \end{subfigure}
    \hfill
    \begin{subfigure}[b]{0.49\linewidth}
        \includegraphics[width=\linewidth]{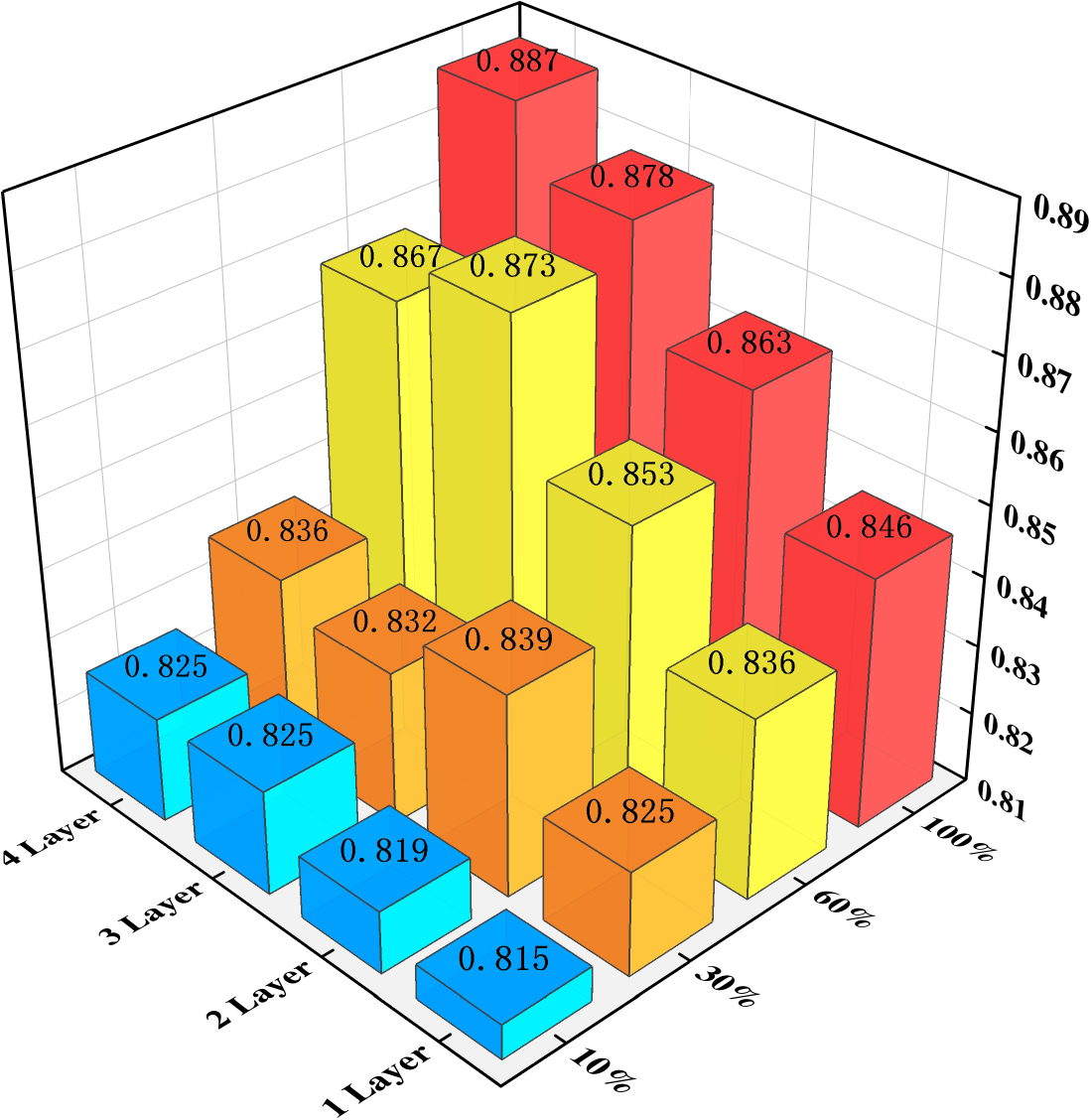}
        \caption{SelfRegulationSCP1}

    \end{subfigure}
    \caption{Results on AWR and SRSCP under different data sizes and layer numbers.}
    \label{fig_scaling}
\end{figure}

We compare the scalability of FAIM by evaluating its performance under various data scales and layer depths. 
Specifically, we conduct experiments using two different UEA datasets of ArticularyWordRecognition (AWR) and SelfRegulationSCP1 (SRSCP), as shown in Figure~\ref{fig_scaling}. 
On smaller data scales, FAIM maintains high accuracy on AWR. As the dataset size increases, the Acc also improves, reaching 0.993 with 100\% of the data. 
This indicates that FAIM can effectively leverage more data samples to enhance its performance, and SRSCP exhibits a similar trend. 
These results further validate the adaptability and scalability of FAIM across different data scales.
Regarding the number of layers, when the layer depth increases from 1 to 4, FAIM shows stable or slightly improved performance on smaller datasets. 
However, for larger datasets, performance slightly decreases as the number of layers increases, which may imply a higher risk of overfitting with greater model complexity. 
Overall, FAIM demonstrates strong robustness on small datasets by maintaining a high performance level, and achieves increased accuracy on larger datasets through effective utilization of additional information. 

\begin{figure*}[hb]
    \begin{subfigure}[b]{0.3\linewidth}
        \includegraphics[width=\linewidth]{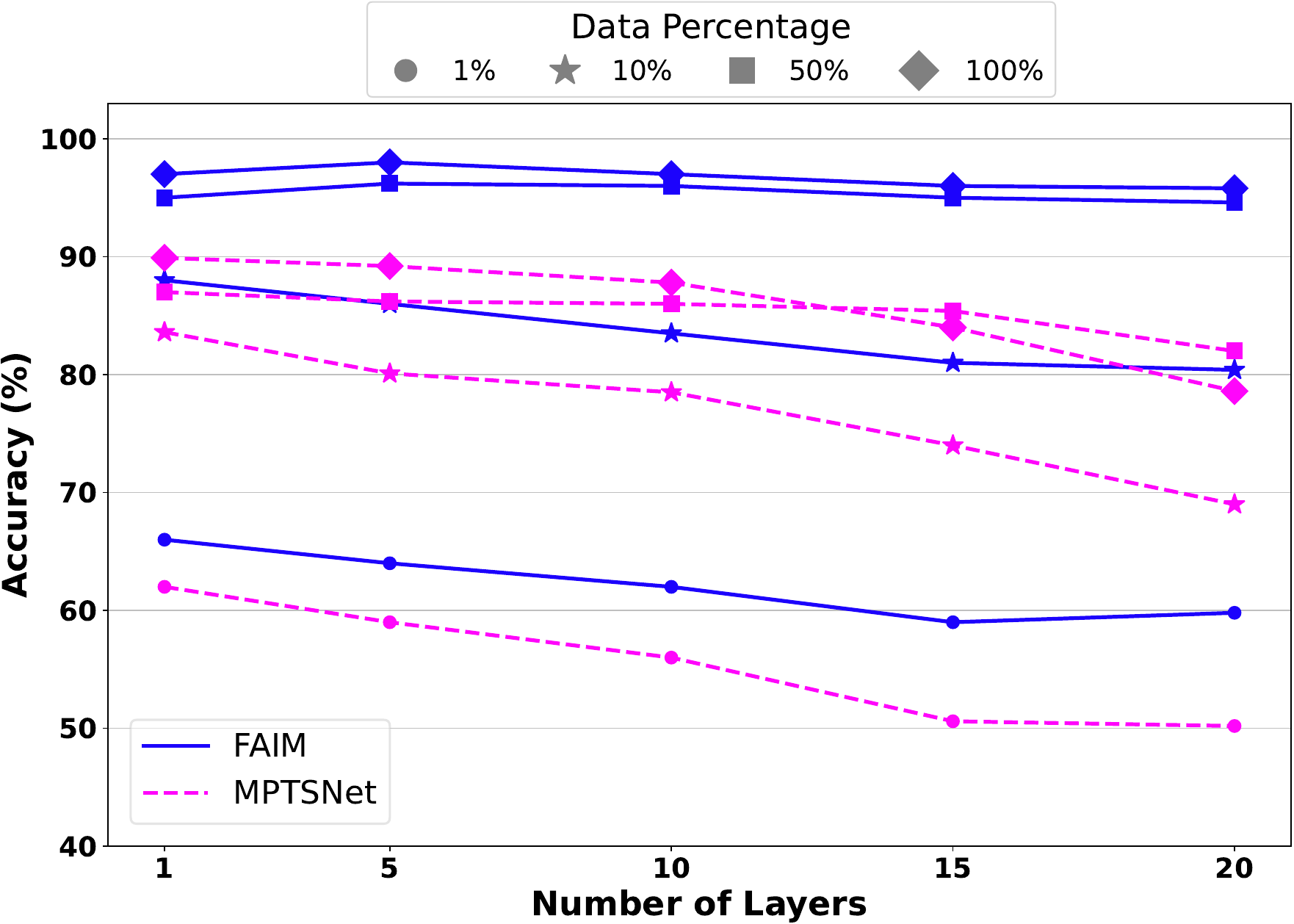}
        \label{fig:sub1}
    \end{subfigure}
    \hfill
    \begin{subfigure}[b]{0.3\linewidth}
        \includegraphics[width=\linewidth]{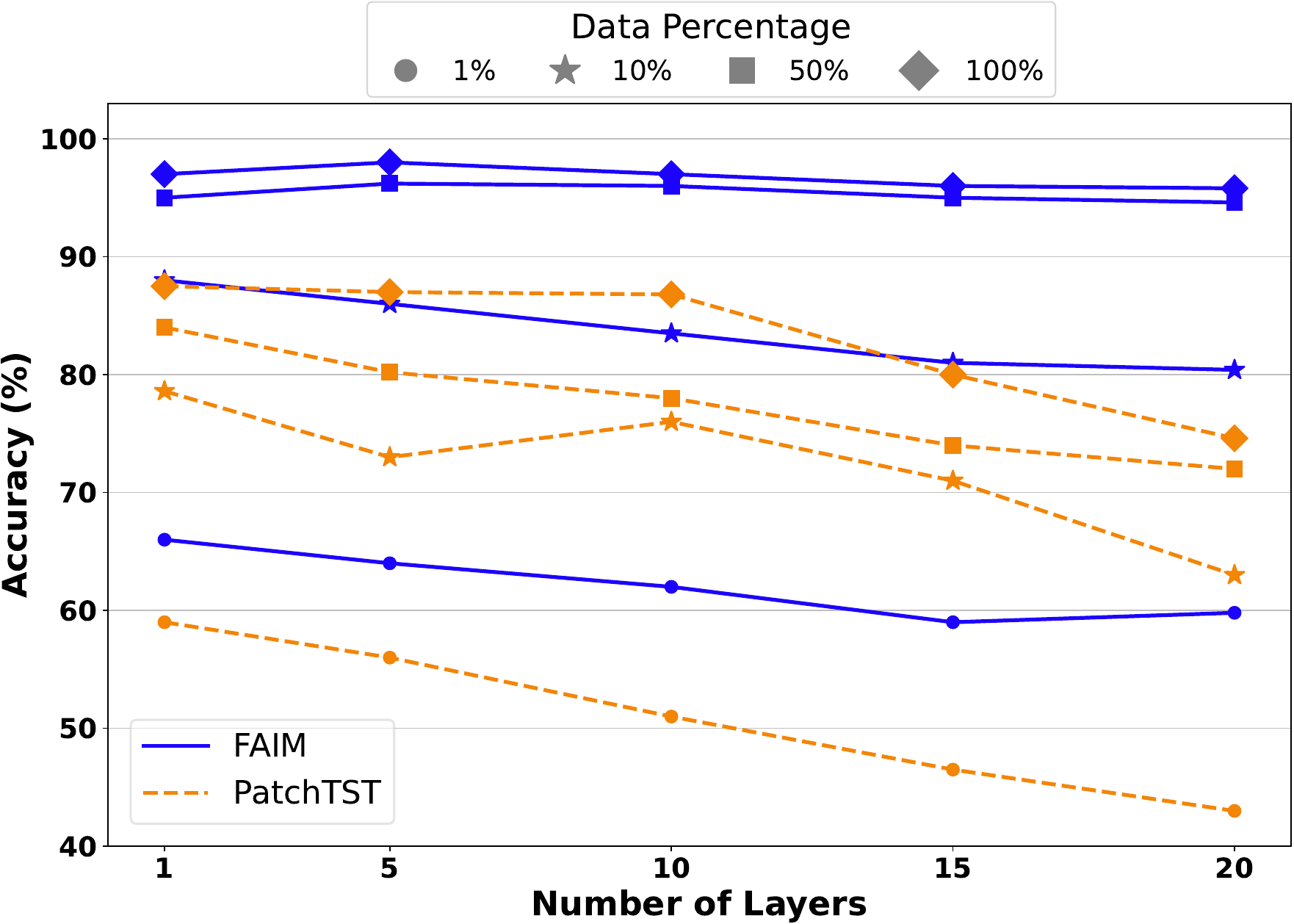}
        \label{fig:sub2}
    \end{subfigure}
    \hfill
    \begin{subfigure}[b]{0.3\linewidth}
        \includegraphics[width=\linewidth]{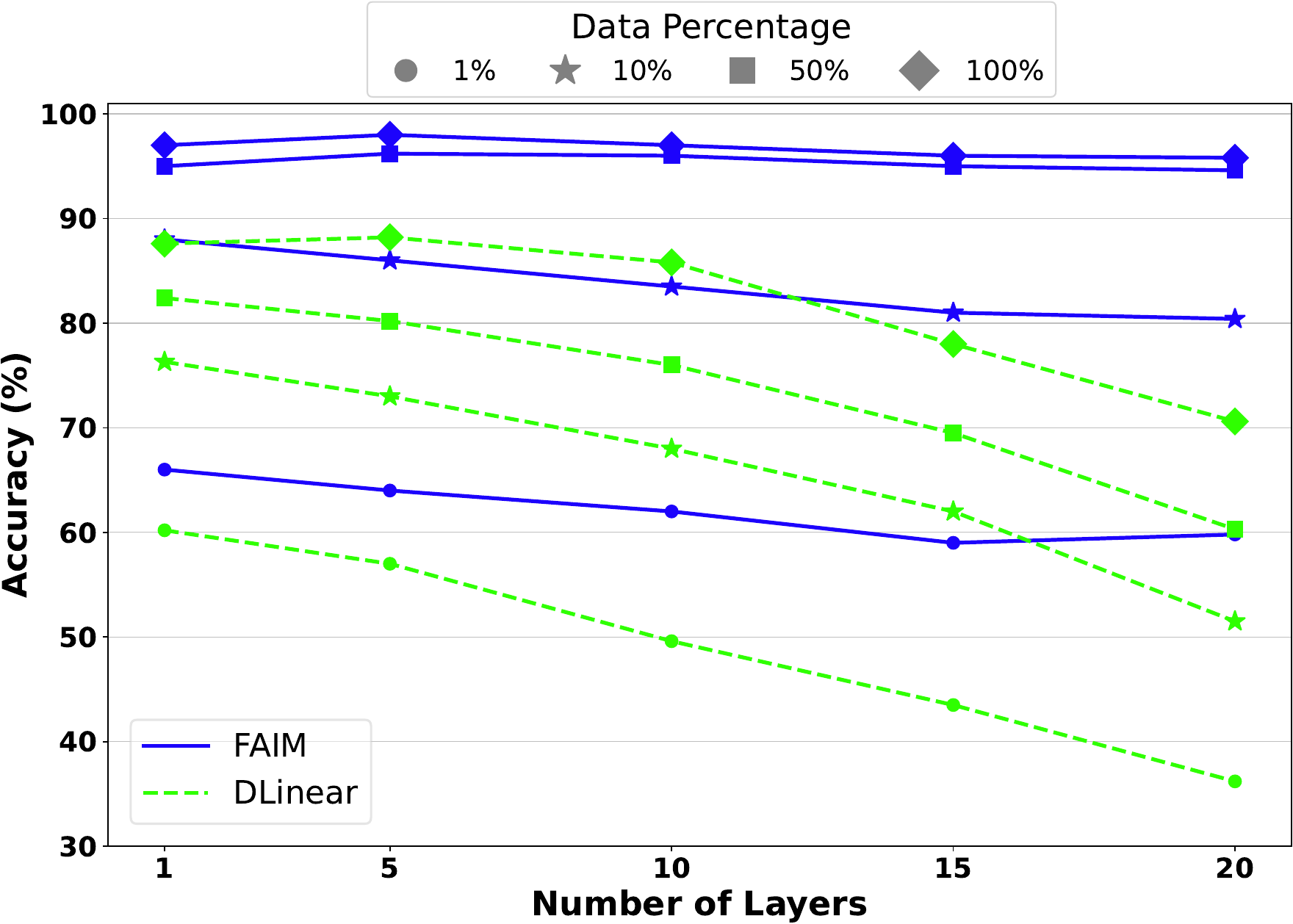}
        \label{fig:sub3}
    \end{subfigure}
    
    \caption{Comparison of the accuracy differences between TSLanet, MPTSNet, PatchTST, and MLP for different data percentages of the uWaveGestureLibraryAll dataset, while changing the number of layers.}
    \label{fig_scale2}
\end{figure*}

We provide a comparison of FAIM with MPTSNet (based on CNN), PatchTST (based on Transformer), and DLinear (based on MLP), observing their performance under different data sizes and layer counts. 
Specifically, we implement various data sizes from the uWaveGestureLibraryAll (UCR) dataset, as shown in Figure \ref{fig_scale2}. 
Notably, at small data sizes, FAIM maintains a consistent level of accuracy, with a slight decline as the number of layers increases. In contrast, other methods exhibit a significant decline in accuracy as the number of layers increases, suggesting potential overfitting issues.

As the dataset size grows, FAIM's performance remains stable, with accuracy varying slightly as the number of layers increases. 
This stability contrasts with the performance of PatchTST and DLinear, which often decline significantly at higher layer counts. 
The trend observed in PatchTST may be attributed to its inherent design, which could lead to diminishing returns or optimization challenges as model depth increases. 
The trend observed in DLinear may be attributed to its MLP structure, which can lead to the vanishing gradient problem as model depth increases. 
MPTSNet performs relatively well, and we believe this is due to its effective integration of frequency modeling. 
Finally, we note that FAIM can effectively utilize larger dataset samples, as its performance improves with increasing number of layers, indicating its ability to leverage more data to enhance accuracy.

\subsubsection{Training Speed and Memory}

\begin{figure}[t]
    \includegraphics[width=0.5\textwidth]{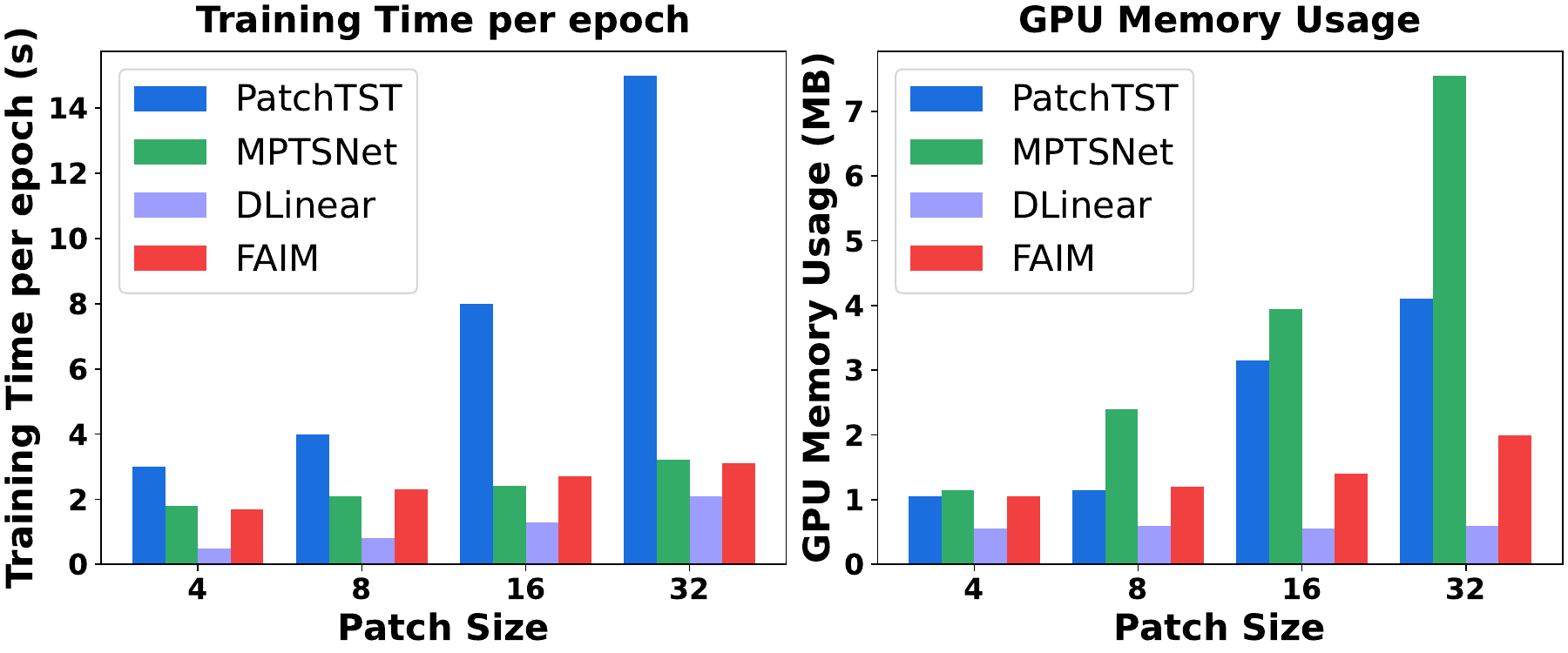}
    \caption{
    Parameters of different patch lengths and GPU memory on SRSCP dataset.
    }
    \label{fig_gpu}
\end{figure}

Considering that hyperparameters set during different training processes and patch lengths may affect the number of model parameters and running speed, we unify the training hyperparameters to make a fairer comparison of parameter count and training speed across different models. Specifically, we select three representative models: PatchTST \cite{Yuqietal-2023-PatchTST}, DLinear \cite{zeng2023transformers}, and MPTSNet \cite{mu2025mptsnet}, and compare them with our proposed model under the same settings. 
Results for different patch lengths on SRSCP are shown in Figure~\ref{fig_gpu}. As observed in the figure, the training time for PatchTST increases with the input length, while the memory consumption of MPTSNet increases significantly. Although DLinear has the fastest training speed, its accuracy is inferior (see the results in the appendix). In contrast, the training time and memory usage of the proposed FAIM increase slowly as the input length grows, demonstrating the superiority of the proposed model in terms of both efficiency and effectiveness.

\subsubsection{Sensitivity Analysis}

\begin{table}[h]
    \centering
\resizebox{1.0\linewidth}{!}{
\begin{tabular}{cc|ccc|ccc}
\toprule
\multirow{2}{*}{\textbf{Conv\_1}} & \multirow{2}{*}{\textbf{Conv\_2}}
 & \multicolumn{3}{c|}{\textbf{FordA (UCR)}} 
 & \multicolumn{3}{c}{\textbf{Epilepsy (UEA)}} \\
 &  & 4 & 8 & 16 & 4 & 8 & 16 \\
\midrule
\multirow{3}{*}{1} 
 & 3 & 0.903 & 0.923 & 0.914 & 0.835 & 0.858 & 0.858 \\
 & 4 & 0.920 & 0.917 & 0.923 & 0.845 & 0.865 & 0.860 \\
 & 5 & 0.919 & 0.920 & 0.917 & 0.845 & 0.866 & 0.862 \\
\midrule
\multirow{3}{*}{2}
 & 3 & 0.937 & 0.933 & 0.922 & 0.964 & 0.985 & 0.981 \\
 & 4 & 0.936 & \textbf{0.944} & 0.938 & 0.975 & \textbf{0.986} & 0.980 \\
 & 5 & 0.941 & 0.930 & 0.931 & 0.970 & \textbf{0.986} & 0.966 \\
\bottomrule
\end{tabular}
}
 \caption{Sensitivity results of kernel size and patch length.}
 \label{table3}
\end{table}

We evaluated the performance of FAIM with three different patch lengths as well as under various convolution kernel sizes (Table \ref{table3}). 
The experimental results show that different patch lengths do not have a particularly significant impact on the results, but FAIM performs more robustly when a moderate patch length is chosen. 
Considering both model efficiency and performance, we set the default patch length to 8. 
Regarding the kernel size in the information interaction stage, when $Conv\_1$ is set to $1 \times 1$, it degenerates to point-to-point matching, which may miss important local structural information and thus leads to inferior performance. 
For $Conv\_2$, we set the convolutional kernel size by default to 4, which achieves a good balance between model performance and computational efficiency. 
Setting the kernel size too large will increase computational overhead and may also introduce irrelevant contextual noise.
We also analyze the sensitivity of the mask ratio parameter settings on the FaceAll and Epilepsy datasets. As shown in Table \ref{table4}, the best performance is achieved when the mask\_ratio is set to 0.4. Proper parameter selection can improve generalization ability and enhance adaptability to unstable and missing data scenarios.

\begin{table}[ht]
\centering
\resizebox{1.0\linewidth}{!}{
\begin{tabular}{l|cccccc}
\toprule
\textbf{Mask Ratio} & 0.25 & 0.3 & 0.35 & 0.4 & 0.45 & 0.5 \\
\midrule
FaceAll (UCR)   & 0.848 & 0.847 & 0.851 & \textbf{0.856} & 0.853 & 0.843 \\
Epilepsy (UEA)  & 0.983 & \textbf{0.986} & 0.985 & \textbf{0.986} & 0.984 & 0.980 \\
\bottomrule
\end{tabular}
}
\caption{Sensitivity results of mask ratio.}
\label{table4}
\end{table}

\subsubsection{Visualization}
\begin{figure*}[hb]
    \centering
    \begin{subfigure}[b]{0.23\linewidth}
        \includegraphics[width=\linewidth]{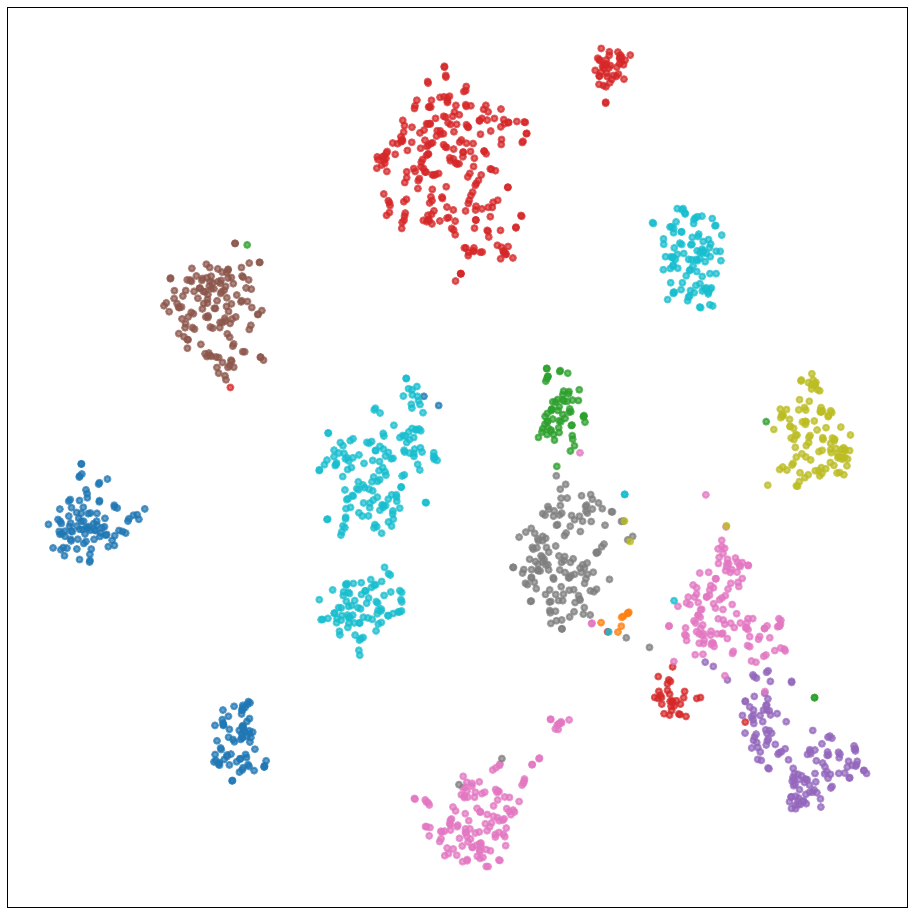}
        \caption{FAIM}
    \end{subfigure}
    \hfill
    \begin{subfigure}[b]{0.23\linewidth}
        \includegraphics[width=\linewidth]{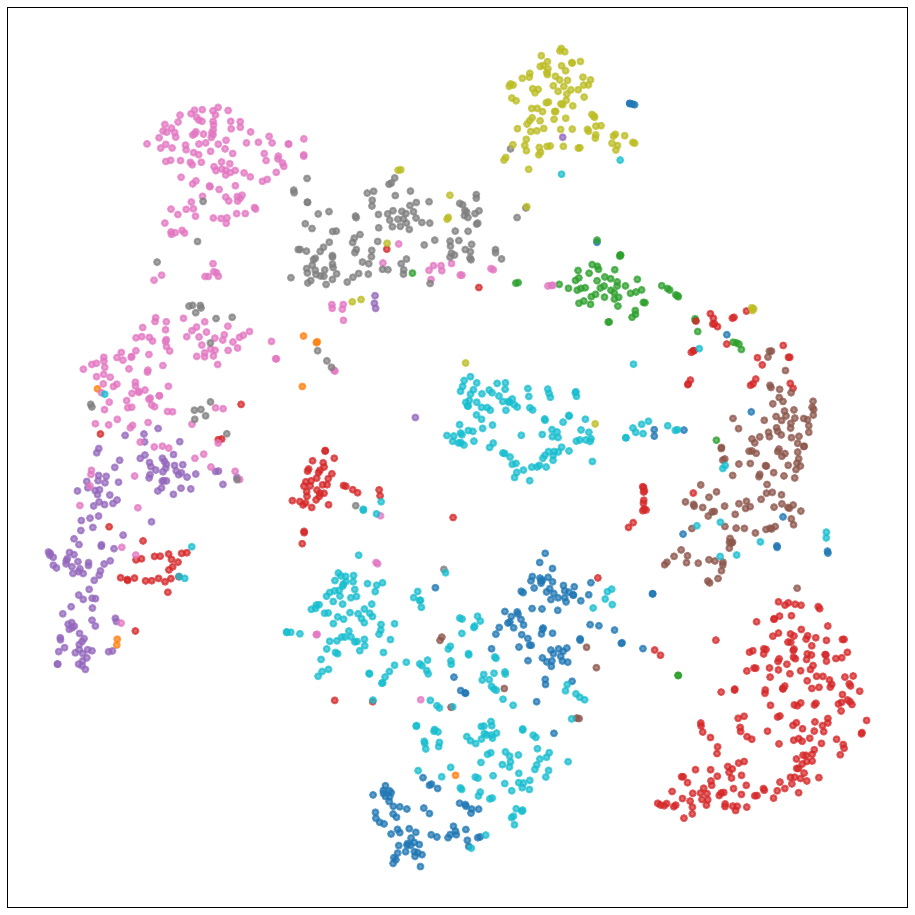}
        \caption{PatchTST}
    \end{subfigure}
    \hfill
    \begin{subfigure}[b]{0.23\linewidth}
        \includegraphics[width=\linewidth]{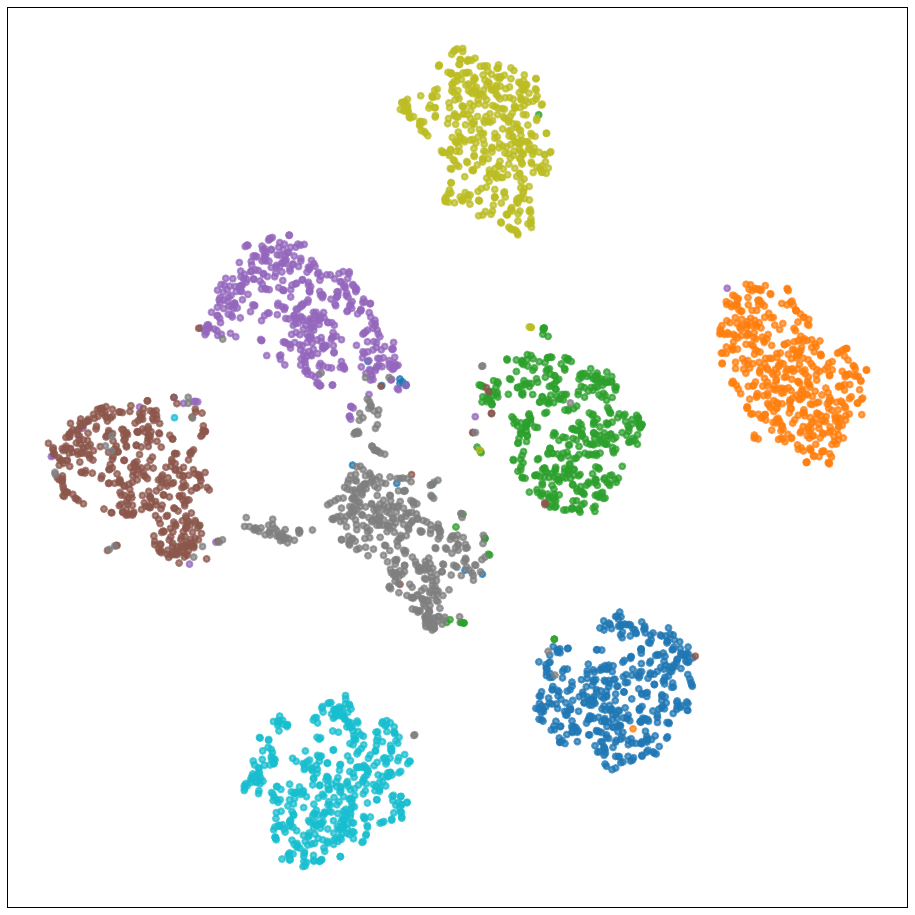}
        \caption{FAIM}
    \end{subfigure}    
    \hfill
    \begin{subfigure}[b]{0.23\linewidth}
        \includegraphics[width=\linewidth]{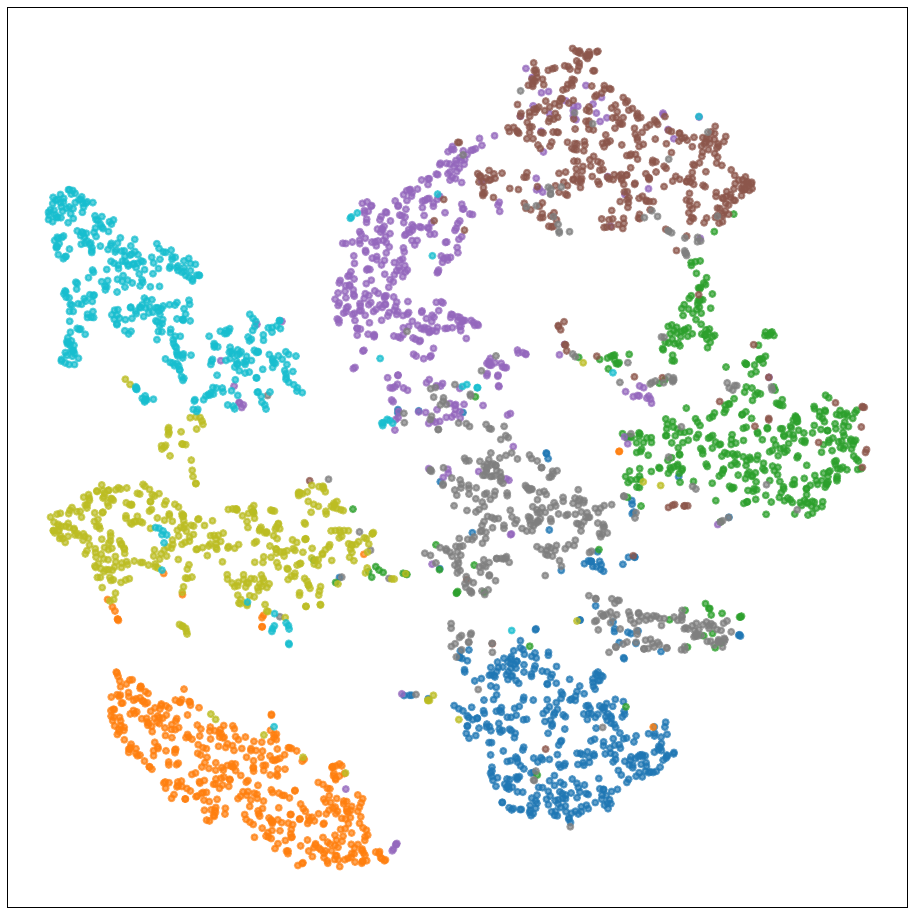}
        \caption{PatchTST}
    \end{subfigure}

    \caption{T-SNE visualization of two methods on FaceAll (UCR) in (a)\&(b) and UWaveGestureLibrary (UEA) in (c)\&(d).}
    \label{fig_tsne}
\end{figure*}

\begin{figure*}[hb]
    \centering
    \begin{subfigure}[b]{0.24\linewidth}
        \includegraphics[width=\linewidth]{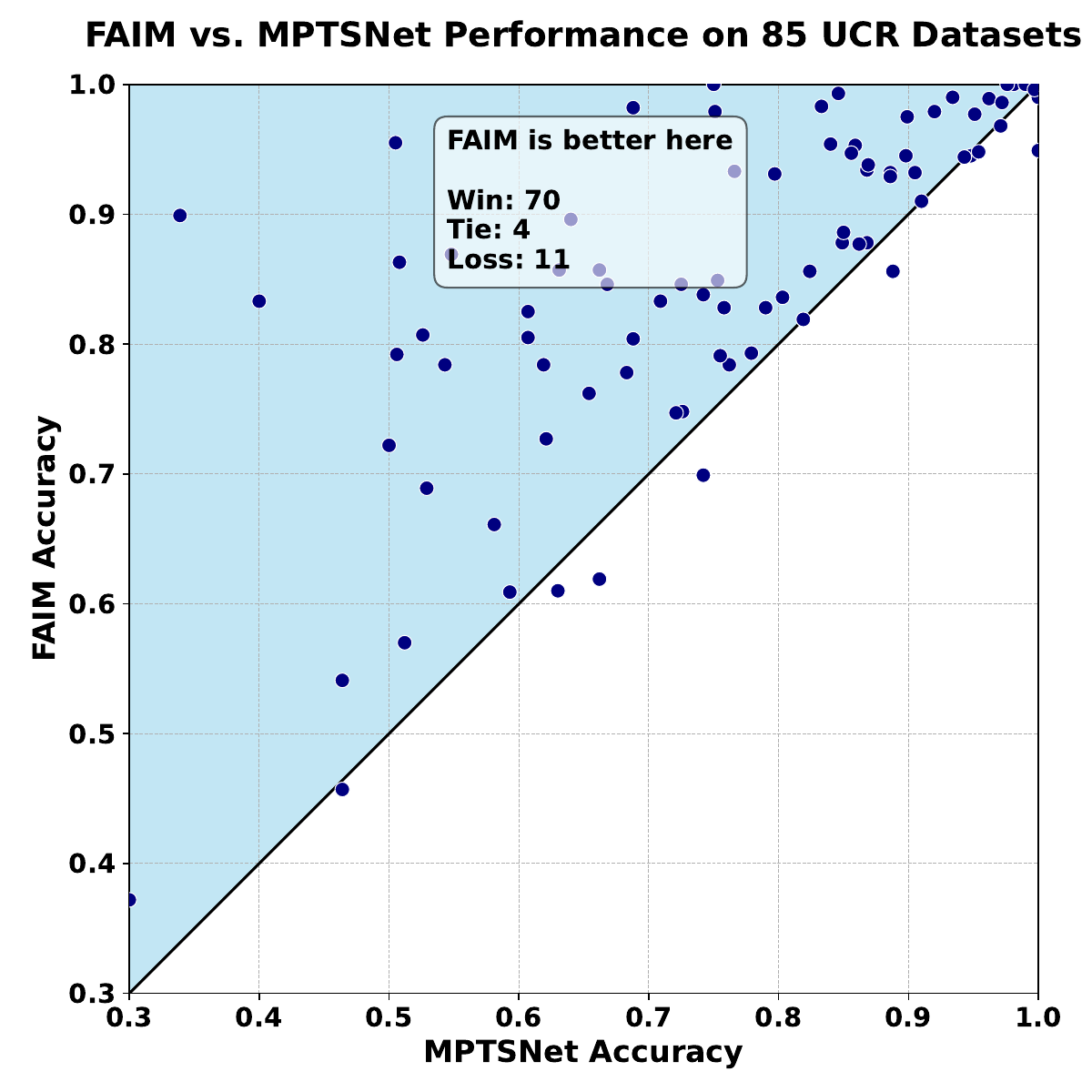}
    \end{subfigure}
    \hfill
    \begin{subfigure}[b]{0.24\linewidth}
        \includegraphics[width=\linewidth]{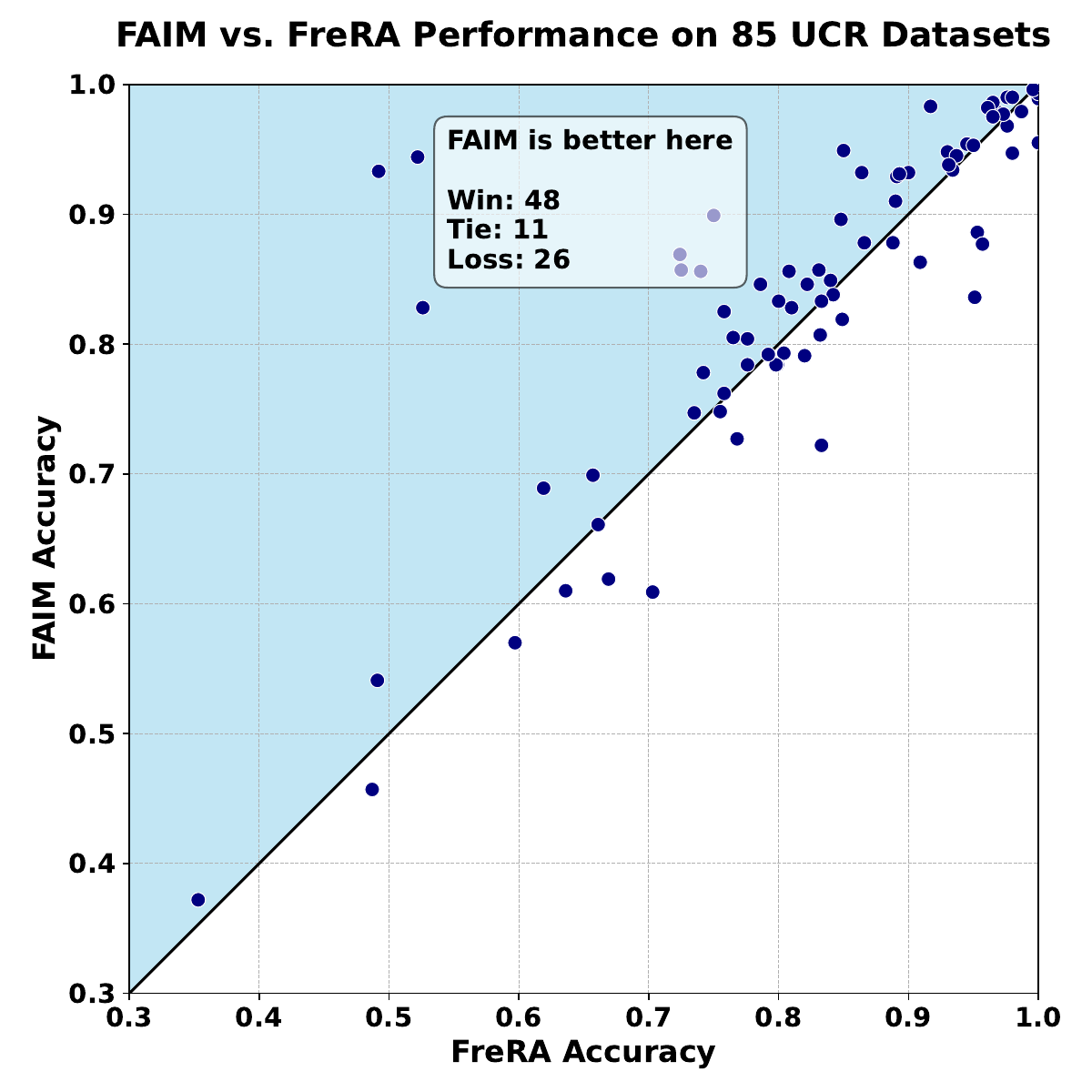}
    \end{subfigure}
    \hfill
    \begin{subfigure}[b]{0.24\linewidth}
        \includegraphics[width=\linewidth]{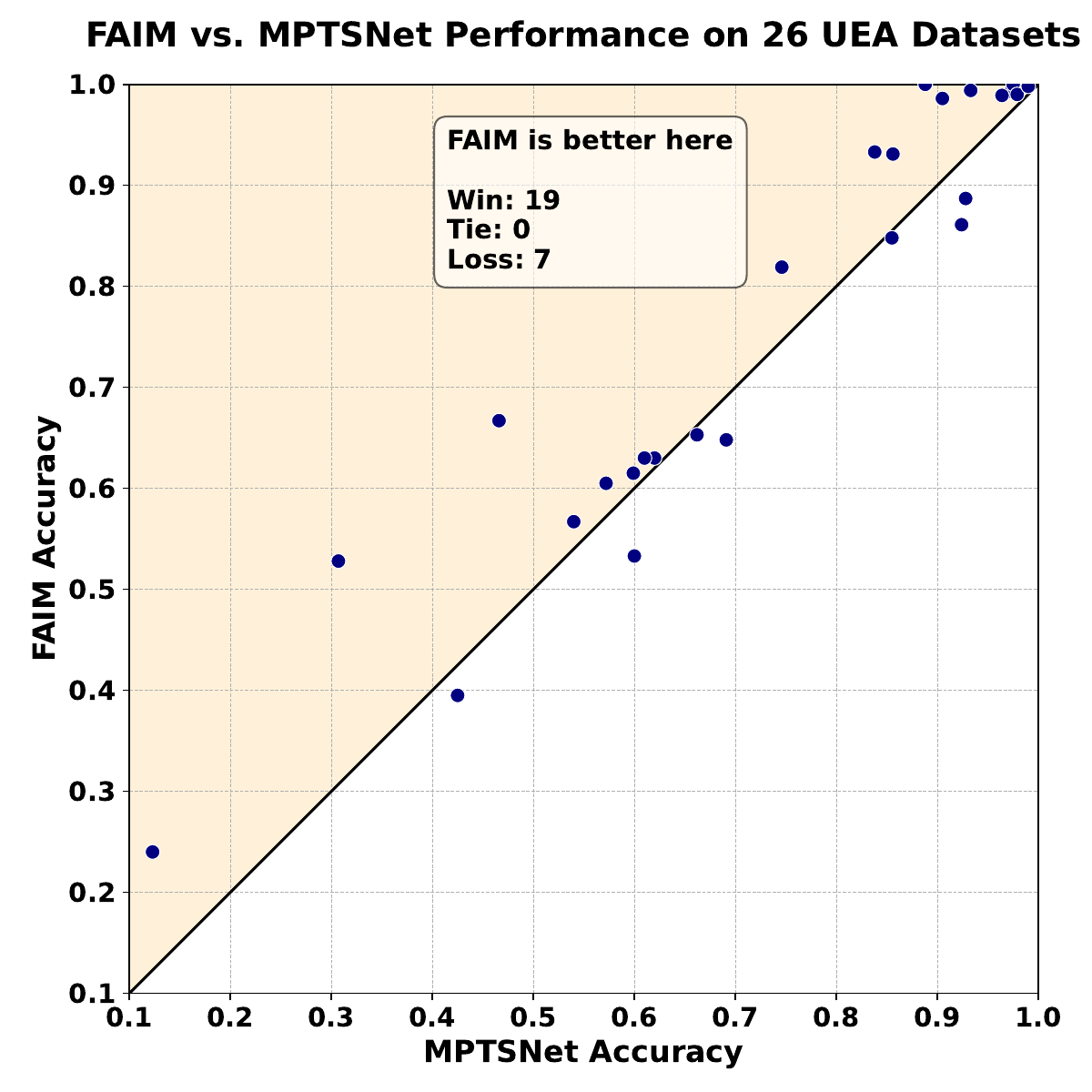}
    \end{subfigure}    
    \hfill
    \begin{subfigure}[b]{0.24\linewidth}
        \includegraphics[width=\linewidth]{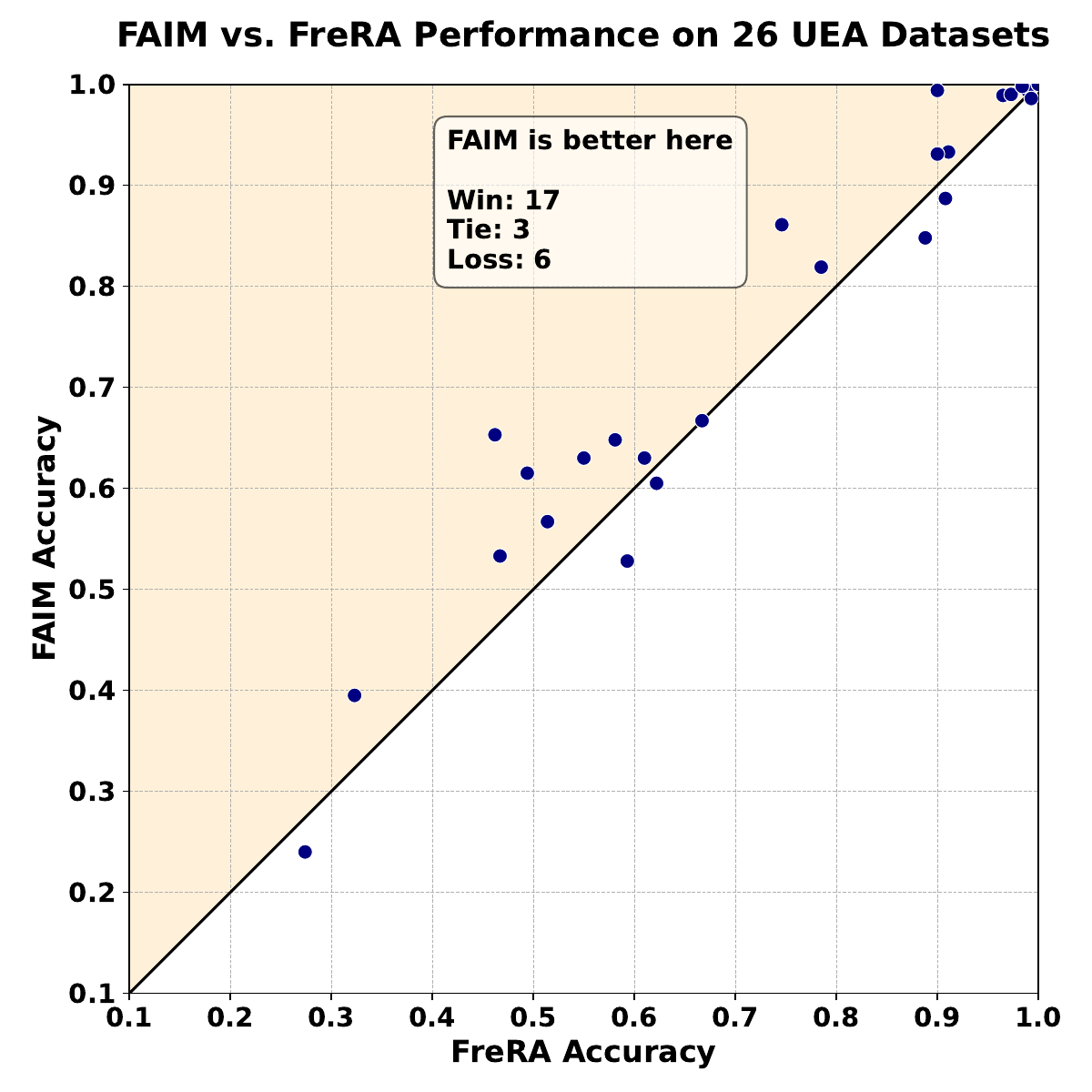}
    \end{subfigure}

    \caption{Scatter plot comparing FAIM with current SOTA methods.}
    \label{fig_compare}
\end{figure*}

To further demonstrate the representational learning capability of FAIM, we project the learned representations onto two-dimensional space for visualization using the t-SNE algorithm \cite{maaten2008visualizing}. 
As shown in Figure~\ref{fig_tsne}, the two plots on the left present results on FaceAll, and the two plots on the right correspond to UWaveGestureLibrary. The results indicate that FAIM can bring samples of the same class closer together, which suggests it can effectively represent class prototypes. 
Such a compact clustering structure means that FAIM better captures the intrinsic features of the data, leading to higher classification accuracy. In contrast, the embeddings produced by PatchTST are more scattered, with less clearly defined boundaries between different classes compared to FAIM. 
This results in inferior classification performance, as the similarity among samples is not fully exploited.

To provide a more intuitive dataset-by-dataset performance comparison, we present scatter plots of FAIM against two strong baselines, MPTSNet and FreRA, in Figure \ref{fig_compare}. It is clearly observable that on both the 85 UCR univariate and 26 UEA multivariate datasets, the vast majority of points are located above the diagonal line. This indicates that FAIM outperforms these competitors in the majority of individual test cases; for instance, on the UCR datasets, FAIM wins against MPTSNet and FreRA on 59 and 48 datasets, respectively. This consistent advantage visually confirms the superior and robust performance of FAIM across diverse data distributions.

\section{Conclusion}
We propose FAIM, a novel lightweight time series classification framework that innovatively combines Mamba with adaptive Fourier filters. It introduces dual-interaction Mamba and employs dropout regularization to enhance parameter selectivity and prevent overfitting. Additionally, it integrates adaptive global and local Fourier filters in the frequency domain, where the local neural operator utilizes learnable thresholds to filter out noise. Experiments demonstrate that FAIM achieves a state-of-the-art balance between performance and efficiency in TSC, showing outstanding results particularly under noisy conditions and across varying data scales. Complexity analysis further confirms the superiority of FAIM, as it substantially reduces computational costs. In future work, we plan to explore more flexible feature interaction mechanisms, investigate model compression and acceleration for large-scale deployment, and extend FAIM to support multi-modal or irregularly sampled time series data to further enhance its practicality and robustness.


\bibliographystyle{IEEEtran}
\bibliography{TKDE_main}

\end{document}